\documentclass[11pt]{article}

\usepackage[T1]{fontenc}
\usepackage{lmodern}

\usepackage[a4paper, margin=1in]{geometry}
\usepackage[expansion=false]{microtype}
\DisableLigatures[f]{family=sf*}
\usepackage{hyphenat}
\usepackage{booktabs}
\usepackage[colorlinks,linkcolor={blue!60!black},citecolor={blue!60!black},urlcolor={blue!70!black}]{hyperref}
\usepackage{graphicx}
\usepackage{xcolor}
\usepackage{amsmath}
\usepackage{enumitem}
\usepackage{listings}
\usepackage{url}
\usepackage{xurl}   
\usepackage{float}
\usepackage{tabularx}
\usepackage{multirow}
\usepackage{fontawesome5}
\usepackage{tabularray}
\UseTblrLibrary{booktabs}
\usepackage{tcolorbox}
\tcbuselibrary{listings, skins, breakable}
\usepackage{fancyhdr}
\usepackage{titling}
\usepackage{titlesec}
\titleformat{\section}{\Large\sffamily\bfseries}{\thesection}{1em}{}
\titleformat{\subsection}{\large\sffamily\bfseries}{\thesubsection}{1em}{}
\titleformat{\subsubsection}{\normalsize\sffamily\bfseries}{\thesubsubsection}{1em}{}
\titleformat{\paragraph}[runin]{\sffamily\bfseries}{\theparagraph}{1em}{}
\usepackage[font=small]{caption}
\DeclareCaptionFormat{sflabel}{{\sffamily\bfseries #1#2}#3}
\usepackage[noabbrev,nameinlink]{cleveref}
\usepackage[section]{placeins}   
\let\oldsubsection\subsection
\renewcommand{\subsection}{\FloatBarrier\oldsubsection}
\let\oldsubsubsection\subsubsection
\renewcommand{\subsubsection}{\FloatBarrier\oldsubsubsection}

\definecolor{boxBg}{HTML}{FAFBFE}
\definecolor{boxFrame}{HTML}{D6DCE8}
\definecolor{headerBg}{HTML}{DEE6F2}
\definecolor{accentBlue}{HTML}{3B7DD8}
\definecolor{labelGray}{HTML}{6B7B95}
\definecolor{codeBg}{HTML}{EAECF3}
\definecolor{tagGreen}{HTML}{1A9A5C}
\definecolor{tagGreenBg}{HTML}{E5F5EB}
\definecolor{tagGrayBg}{HTML}{E2E5EC}
\definecolor{tableHeader}{HTML}{2E5E8C}
\definecolor{tableStripe}{HTML}{F2F6FA}
\definecolor{tableRule}{HTML}{2E5E8C}

\newtcolorbox{skillbox}[1][]{%
  enhanced, breakable,
  colback    = boxBg,
  colframe   = boxFrame,
  coltitle   = accentBlue,
  fonttitle  = \sffamily\bfseries\small,
  arc        = 5pt,
  boxrule    = 0.5pt,
  left       = 10pt,
  right      = 10pt,
  top        = 6pt,
  bottom     = 6pt,
  toptitle   = 5pt,
  bottomtitle= 5pt,
  title      = {#1},
  colbacktitle = headerBg,
}

\newtcblisting{codebox}[1][]{%
  enhanced, breakable,
  colback    = codeBg,
  colframe   = boxFrame,
  coltitle   = accentBlue,
  fonttitle  = \sffamily\bfseries\small,
  arc        = 5pt,
  boxrule    = 0.5pt,
  left       = 8pt,
  right      = 8pt,
  top        = 6pt,
  bottom     = 6pt,
  toptitle   = 5pt,
  bottomtitle= 5pt,
  colbacktitle = headerBg,
  listing only,
  listing options = {
    basicstyle  = \ttfamily\footnotesize,
    breaklines  = true,
    columns     = fullflexible,
    keepspaces  = true,
  },
  #1
}

\newtcolorbox{innercode}{%
  enhanced,
  colback  = codeBg,
  colframe = boxFrame,
  arc      = 3pt,
  boxrule  = 0.4pt,
  left     = 6pt,
  right    = 6pt,
  top      = 4pt,
  bottom   = 4pt,
}

\usepackage[backend=biber,style=numeric,sorting=none]{biblatex}
\pretitle{\begin{center}\LARGE\bfseries}
\posttitle{\par\end{center}\vspace{0.35em}}
\preauthor{\begin{center}\large}
\postauthor{\par\end{center}\vspace{-0.4em}}
\predate{}\postdate{}

\title{SkillCenter: A Large-Scale Source-Grounded Skill Library for Autonomous AI Agents}
\author{%
Tianming Sha\textsuperscript{1}\quad Yue Zhao\textsuperscript{2}\quad Lichao Sun\textsuperscript{3}\quad Yushun Dong\textsuperscript{4,\,$\dagger$}\\[0.45em]
{\small\textsuperscript{1}Stony Brook University\quad\textsuperscript{2}University of Southern California\\[0.12em]
\textsuperscript{3}Lehigh University\quad\textsuperscript{4}Florida State University}\\[0.4em]
{\footnotesize\ttfamily tianming.sha@stonybrook.edu\quad yzhao010@usc.edu\quad lis221@lehigh.edu\quad yd24f@fsu.edu}\\[0.35em]
{\small $\dagger$\,Corresponding author}%
}
\date{}


\begin{document}
\raggedbottom

\fancypagestyle{firstpage}{%
  \fancyhf{}%
  \renewcommand{\headrulewidth}{0pt}%
  \renewcommand{\footrulewidth}{0pt}%
}

\thispagestyle{firstpage}

\begin{tcolorbox}[
  enhanced,
  frame hidden,
  colback    = boxBg,
  arc        = 12pt,
  left       = 18pt,
  right      = 18pt,
  top        = 14pt,
  bottom     = 14pt,
  before skip= 0pt,
  after skip = 1.2em,
  grow to left by  = 1pt,
  grow to right by = 1pt,
  breakable,
  overlay unbroken and first = {\node[anchor=north west,
                  at=(frame.north west),
                  xshift=14pt, yshift=-8pt]
                  {\includegraphics[height=0.9cm]{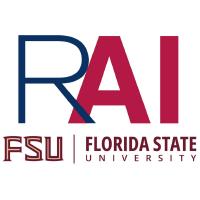}};},
]
\setlength{\parindent}{0pt}

\vspace{0.8cm}
{\centering\LARGE\sffamily\bfseries \thetitle\par}
\vspace{0.6em}

{\centering\large \theauthor\par}
\vspace{0.6em}

{\centering\includegraphics[width=0.81\linewidth]{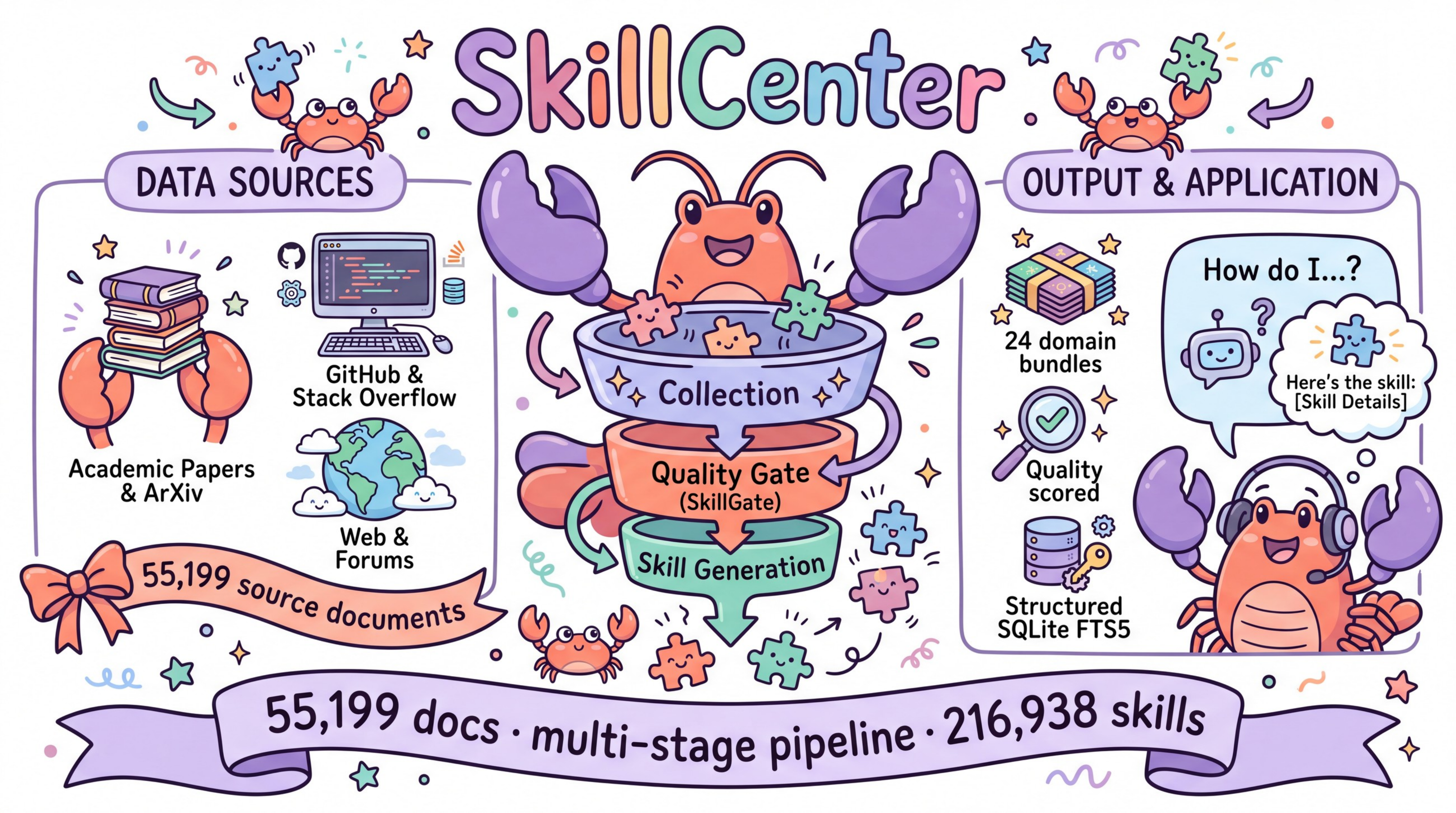}\par}
\vspace{0.5em}

{\sffamily\bfseries\color{accentBlue} Abstract}\par
\smallskip
\small
Autonomous AI agents can execute complex tasks with limited human review, yet they often lack the grounded operational knowledge to make their outputs not just executable but correct, secure, and maintainable. We introduce \textbf{SkillCenter}, to our knowledge the largest open skill library for agents by total count: 216,938 structured skills across 24 domain bundles. A SkillGate-filtered pipeline contributes 114,565 source-grounded skills from peer-reviewed journals, ArXiv, and over 24,000 technical sources, integrated with 102,373 community skills from GitHub and the ClawHub marketplace. We present the end-to-end framework that builds the pipeline subset: multi-source acquisition, an LLM-based quality gate (SkillGate), template-driven generation, iterative source-grounding, and quality-controlled publishing. Source grounding is a traceability guarantee: each retained claim maps to an exact quotation in its source. All skills ship as offline-searchable SQLite FTS5 bundles.

\vspace{0.55em}
\noindent\textcolor{boxFrame}{\rule{\linewidth}{0.4pt}}
\vspace{0.4em}

{\small
\sffamily\bfseries\color{labelGray} GitHub:\;\normalfont\url{https://github.com/LabRAI/SkillCenter}\par
\vspace{2pt}
\sffamily\bfseries\color{labelGray} Hugging Face:\;\normalfont\url{https://huggingface.co/datasets/Tommysha/skillcenter-bundles}\par
}

\end{tcolorbox}

\newpage
\section{Introduction}

When a developer asks an AI agent to ``implement the authentication module,'' the agent can usually produce a result that runs. Whether that result is correct, secure, and maintainable is a separate matter. As large language models (LLMs) and autonomous agents take over tasks that practitioners once performed by hand, a quality gap emerges: agents gain autonomy faster than they gain the operational judgment that human expertise once supplied. We argue that \textbf{skills}, defined as structured, retrievable, source-grounded units of operational knowledge, can serve as infrastructure for narrowing this gap. This paper introduces \textbf{SkillCenter}, a library of 216,938 such skills across 24 domain bundles, together with an automated framework for producing, validating, and distributing them at scale.

\subsection{Vibe Coding and Vibe Research}

In February 2025, Andrej Karpathy coined the term vibe coding~\cite{karpathy2025vibe} for a workflow in which a developer states intent in natural language, an LLM writes the code, and the developer accepts or rejects the output based on whether the program appears to work, without line-by-line review. Code-generation tools such as GitHub Copilot had existed for years; the term named a newer disposition, the willingness to stop reading the generated code. It gained currency because it described a practice that was already widespread.

The idea migrated quickly. OpenAI researchers Jakub Pachocki and Mark Chen applied the same logic to scientific inquiry under the label vibe research~\cite{viberesearch2026}: let AI agents handle literature review, data analysis, and even preliminary drafting while the human researcher steers. The researcher stops being the person who pipettes the reagent and becomes the person who decides which reagent to pipette.

These methodologies are also drawing scholarly attention; the VibeX 2026 workshop at EASE 2026 is one venue dedicated to studying them~\cite{vibex2026}. Industry is moving in parallel: a collaboration between OpenAI and Ginkgo Bioworks paired a GPT-5 model with an autonomous laboratory to design and run biological experiments at scale~\cite{ginkgo2026}.

In effect, the human role is shifting from executing tasks to directing them.

\subsection{The Autonomy Spectrum}

The transition to agent-executed work has not happened in a single leap. We describe it as a four-stage autonomy spectrum:

\begin{enumerate}[leftmargin=2em]
    \item \textbf{Human-only.} The practitioner writes all code, runs all experiments, and reviews all outputs. The machine is a passive execution environment.
    \item \textbf{LLM-assisted.} Tools such as GitHub Copilot, ChatGPT, and similar systems suggest completions, answer questions, and draft short passages. The human retains full editorial control and reviews every suggestion before acceptance.
    \item \textbf{Agent-executed.} Integrated development agents (Claude Code, Cursor, Windsurf, and their counterparts in scientific domains~\cite{agenticbioinfo2026}) receive high-level tasks (``implement the authentication module,'' ``analyze the RNA-seq dataset'') and produce multi-file, multi-step outputs with minimal human intervention. The human reviews selectively.
    \item \textbf{Agentic autonomous.} Multi-agent systems decompose complex goals into sub-tasks, delegate to specialized sub-agents, and orchestrate execution across tools, APIs, and environments~\cite{xi2023agents}. The OpenClaw agent framework~\cite{openclaw2026agent} (an open-source harness for building skill-composing assistants and automation agents) and OPAL~\cite{opal2026} are early examples. In OpenClaw's architecture, agents assemble task-specific behaviors from a registry of discrete skill primitives, making the availability and quality of those primitives a first-order concern. Human involvement at this stage is limited to goal-setting, periodic auditing, and exception handling.
\end{enumerate}

At each stage, the agent's independent decision-making increases while human oversight decreases. When a human writes code, decades of training and institutional knowledge implicitly constrain the solution space; when an agent writes code, those constraints must be made explicit. The autonomy spectrum is therefore also a risk spectrum: the less a human is in the loop, the more the system depends on the quality of the knowledge available to the agent.

\subsection{Execution Versus Correctness}

Modern LLMs are capable code generators. Given a well-formed prompt, an LLM can produce a Python module that parses a dataset, trains a model, and writes results to disk, and the module will often execute without errors on the first attempt. Successful execution, however, is a weak signal of quality: benchmarks of real-world issue resolution show that code which runs can still fail to fix the underlying problem~\cite{jimenez2024swebench}. The module may silently leak memory, use a deprecated API, ignore edge cases in the data, apply an inappropriate statistical test, or introduce a security vulnerability that passes all existing tests. In scientific contexts, a generated analysis may use a valid but suboptimal method or fail to correct for multiple comparisons. Execution success and substantive correctness are distinct properties.

Experience, mentorship, code review, and institutional standards traditionally bridged this gap. In an agent-executed paradigm, these mechanisms are either absent or operate on a different timescale: a code review that takes a senior engineer thirty minutes is at odds with an agent that generates a feature in seconds. The autonomy thus creates a tension. \textbf{The more autonomous the agent, the more it needs grounded knowledge to constrain its behavior.} Yet the very autonomy that makes agents useful removes the human from the position where such knowledge was traditionally applied.

Current mitigation strategies fall into two categories. Human-in-the-loop review is effective but unscalable: the review bottleneck grows proportionally with agent output. Retrieval-augmented generation (RAG) improves factual grounding but operates on raw document chunks that may be noisy, contradictory, or tangential. The agent must still decide what to do with the retrieved passages, a decision that requires the operational judgment it lacks; neither approach scales on its own.

\subsection{Why Skills Are Needed}

We define a \textbf{skill} as a structured, retrievable, source-grounded unit of operational knowledge. Each skill encodes a specific piece of actionable guidance (a best practice, a common pitfall, a validated technique, a configuration recommendation) together with metadata including its domain, evidence provenance, quality score, and applicability conditions. Skills differ from RAG document chunks in several important respects:

\begin{itemize}[leftmargin=2em]
    \item \textbf{Structure.} A skill has a defined schema (title, description, applicability, evidence, score), whereas a RAG chunk is an arbitrary substring of a source document.
    \item \textbf{Quality gating.} Each skill passes through an LLM-based quality gate that scores clarity, accuracy, and actionability (an internal QA signal, not external validation); RAG chunks inherit the quality of their source without independent assessment.
    \item \textbf{Actionability.} A skill is written to be directly applied by an agent during task execution; a RAG chunk provides context that the agent must interpret and operationalize on its own.
    \item \textbf{Scoring.} Skills carry numerical quality scores that allow agents to prioritize high-confidence guidance; RAG systems typically rank by relevance rather than quality.
\end{itemize}

Reusable skill collections already exist in robotics, game AI, LLM agent frameworks, and community marketplaces such as ClawHub~\cite{clawhub2026}, which we survey in \S\ref{sec:related-work}. Existing libraries share three limitations:

\begin{enumerate}[leftmargin=2em]
    \item \textbf{Manual authorship.} Skills are written by hand, making creation slow and expensive. A domain expert may spend hours distilling a single skill from the literature.
    \item \textbf{Limited scale.} The largest publicly available collections contain hundreds to low thousands of skills, far too few to cover the breadth of knowledge that a general-purpose agent requires.
    \item \textbf{Narrow coverage.} Most collections focus on a single domain (e.g., software engineering, robotic manipulation) and do not transfer across the diverse tasks that modern agents are expected to handle.
\end{enumerate}

These libraries are not keeping pace with demand. As agents are deployed in more domains and given more autonomy, the number of situations in which grounded operational knowledge would prevent errors grows combinatorially, and manual curation does not scale to meet it. This motivates an approach that is automated, source-grounded, multi-domain, and able to scale to hundreds of thousands of skills, which is the goal of this work.

\subsection{Our Contributions}

This paper makes three primary contributions:

\begin{enumerate}[leftmargin=2em]
    \item \textbf{The SkillCenter library.} We release 216,938 skills across 24 domain bundles, of which 114,565 are produced by the SkillGate pipeline and 102,373 are integrated from public GitHub repositories and the ClawHub marketplace, exceeding prior open skill collections by more than an order of magnitude in size. Section~\ref{sec:corpus-overview} gives the full breakdown by source and domain.

    \item \textbf{An end-to-end automated framework.} We present a complete, continuously running pipeline for skill production, covering multi-source acquisition, an LLM-based quality gate (SkillGate), template-driven generation, iterative source-grounding, and quality-controlled publishing with full audit trails, that ingests new sources and produces new skills without manual intervention.

    \item \textbf{An offline-searchable distribution system.} We package the corpus as domain-split SQLite FTS5 bundles~\cite{sqlite_fts5} that support offline keyword search over skill titles without network access. An automatic project-type detection mechanism identifies the relevant domains for a given project, and a \texttt{bundle-install --auto} command integrates the appropriate skill bundles into agent workflows. Pre-built bundles are available on Hugging Face\footnote{\url{https://huggingface.co/datasets/Tommysha/skillcenter-bundles}}~\cite{huggingface_datasets}, and the full framework is open-source on GitHub.\footnote{\url{https://github.com/LabRAI/SkillCenter}}
\end{enumerate}

\section{The SkillCenter Corpus}

This section analyzes the corpus: its scale and domain composition (\S\ref{sec:corpus-overview}), academic (\S\ref{sec:academic-sources}) and technical (\S\ref{sec:technical-sources}) sources, the integrated community bundles (\S\ref{sec:community}), deduplication and redundancy (\S\ref{sec:dedup}), quality scores (\S\ref{sec:quality-scores}), and the search and agent integration layer (\S\ref{sec:search-integration}).

\subsection{Corpus Overview}
\label{sec:corpus-overview}

\begin{figure}[t]
\centering
\includegraphics[width=\textwidth]{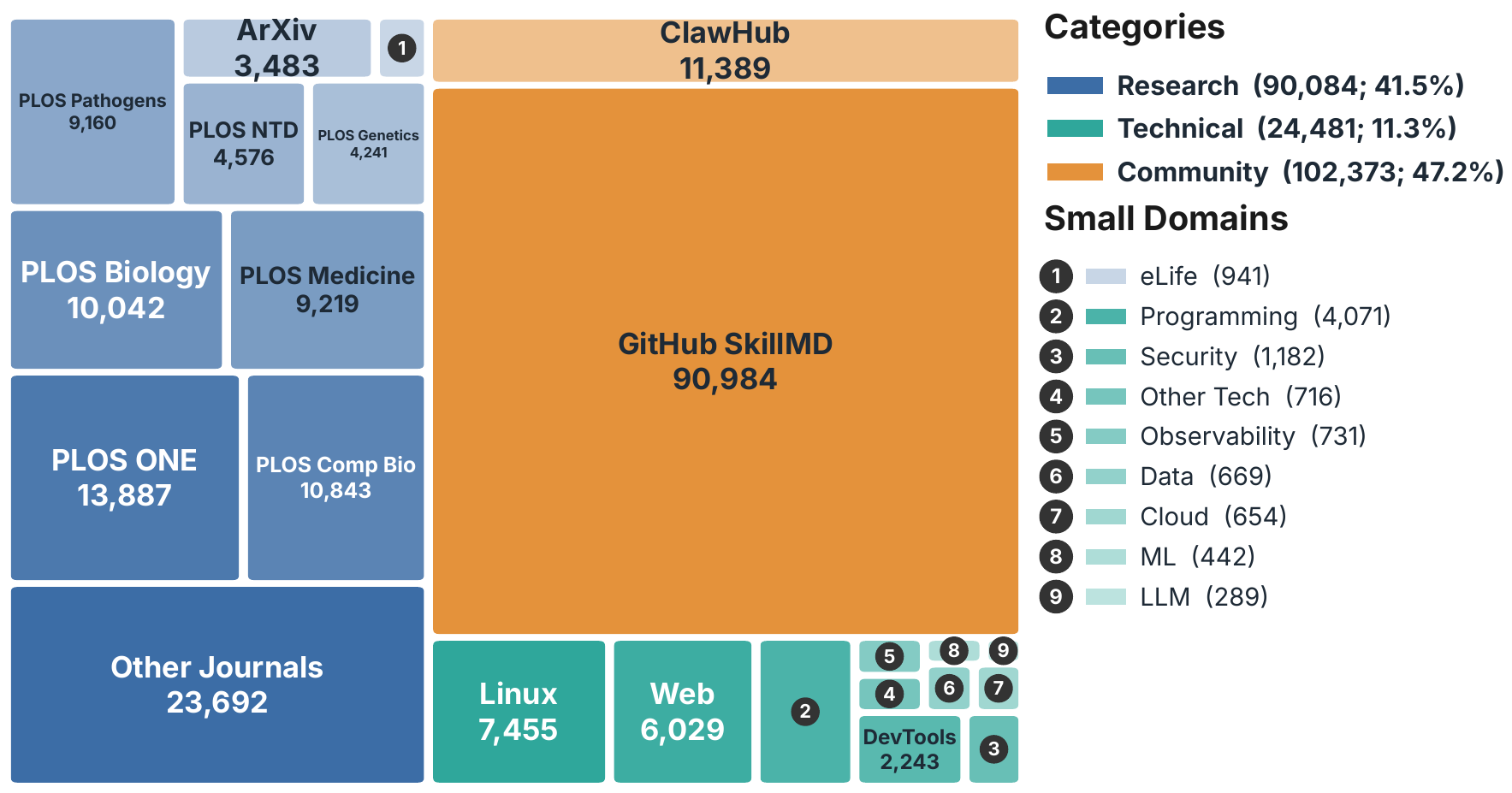}
\caption{Treemap visualization of the SkillCenter library. Area is proportional to skill count. The library comprises pipeline-produced Research (41.5\%) and Technical (11.3\%) bundles plus an integrated Community category of GitHub SkillMD and the ClawHub marketplace (47.2\%).}
\label{fig:treemap}
\end{figure}

The SkillCenter library contains 216,938 published skills organized into 24 domain bundles, drawn from three broad categories: Research (90,084 skills, 41.5\%), Technical (24,481 skills, 11.3\%), and Community (102,373 skills, 47.2\%). The Research and Technical bundles (114,565 skills) are produced by the SkillGate pipeline (\S\ref{sec:framework}) from 55,199 distinct source documents; the Community category integrates two large externally harvested collections: GitHub SkillMD (90,984 \texttt{SKILL.md} files mined from public repositories via code search) and the ClawHub community marketplace (11,389 skills). Figure~\ref{fig:treemap} visualizes the relative scale of these categories, and Table~\ref{tab:corpus-summary} provides the complete per-domain breakdown. All skills are in English. Pipeline-produced skills carry a quality score on a scale from 1 to 5 assigned by GPT-5.2 during the publish gate; their average is 3.91 (the two integrated collections use separate scoring scales and are excluded from this average).

\paragraph{Terminology.}
A few terms recur throughout this report:
\begin{itemize}[leftmargin=1.6em, itemsep=1pt, topsep=2pt]
    \item \textbf{Source}: a raw input document (web page, README, paper, forum post, repository). There are 55,199 sources for the pipeline subset and 94,722 distinct across the full library.
    \item \textbf{Generated skill}: a skill produced by the LLM pipeline before quality filtering. \textbf{Published skill}: one that passes the publish gate and ships in a released bundle.
    \item \textbf{Skill kind}: the facet a skill captures (e.g.\ \texttt{paper.experiment}, \texttt{paper.method}, or the default technical template); one source can yield several kinds.
    \item \textbf{Domain configuration} (13): a framework input spec defining source routing, host lists, and search parameters for one topic domain.
    \item \textbf{Domain bundle} (24): a published SQLite database grouping skills; 21 are produced by the domain configurations and 3 are integrated community bundles.
\end{itemize}
The phrase ``over 24,000 technical sources'' refers to the distinct GitHub, web, and forum source documents that produced the technical skills.

\begin{table}[!htbp]
\centering
\caption{Corpus summary by domain bundle for the full SkillCenter library (216,938 skills). Research and Technical bundles are produced by the SkillGate pipeline (\S\ref{sec:framework}); the Community category integrates two externally harvested collections, GitHub SkillMD (\texttt{SKILL.md} files mined from public repositories via code search) and the ClawHub community marketplace. The former ``Other Journals'' bundle (23,692 skills) is decomposed into its constituent Nature family sub-journals and a residual category. Counts were verified against the published SQLite bundles; the two integrated community collections (GitHub SkillMD and ClawHub) use separate quality scales and are shown without a score on the 1-to-5 scale, so the \textbf{Total} Score is the mean over the pipeline-scored subset (114,565 skills).}
\label{tab:corpus-summary}
\scriptsize
\begin{tblr}{
  colspec    = {lllrrrl},
  row{1}     = {bg=tableHeader, fg=white, font=\bfseries\sffamily},
  row{38}    = {bg=codeBg, font=\bfseries},
  rowsep     = 2pt,
  colsep     = 10pt,
  hline{1,Z} = {0.09em, tableRule},
  hline{9}   = {2-7}{0.03em},
  hline{21}  = {2-7}{0.03em},
  hline{24}  = {0.05em},
  hline{35}  = {0.05em},
  hline{38}  = {0.05em},
}
Category & Family & Domain & Skills & \% & Score & Source \\
\SetCell[r=22]{l} Research
& \SetCell[r=7]{l} PLOS & PLOS ONE & 13,887 & 6.4 & 4.01 & journal \\
& & PLOS Comp Bio & 10,843 & 5.0 & 3.85 & journal \\
& & PLOS Biology & 10,042 & 4.6 & 3.96 & journal \\
& & PLOS Genetics & 4,241 & 2.0 & 3.77 & journal \\
& & PLOS Medicine & 9,219 & 4.2 & 3.94 & journal \\
& & PLOS Pathogens & 9,160 & 4.2 & 3.86 & journal \\
& & PLOS NTD & 4,576 & 2.1 & 3.87 & journal \\
& \SetCell[r=12]{l} Nature & Nat. Neuroscience & 1,303 & 0.6 & 3.81 & journal \\
& & Nat. Cell Biology & 1,163 & 0.5 & 3.77 & journal \\
& & Nat. Methods & 1,160 & 0.5 & 4.05 & journal \\
& & Nat. Immunology & 1,084 & 0.5 & 3.81 & journal \\
& & Nat. Microbiology & 1,073 & 0.5 & 3.84 & journal \\
& & Nature (flagship) & 1,048 & 0.5 & 4.12 & journal \\
& & Scientific Reports & 1,030 & 0.5 & 3.76 & journal \\
& & Nat. Medicine & 998 & 0.5 & 3.90 & journal \\
& & Nat. Human Behaviour & 997 & 0.5 & 3.89 & journal \\
& & Nat. Chemistry & 952 & 0.4 & 3.88 & journal \\
& & Nat. Metabolism & 943 & 0.4 & 3.77 & journal \\
& & 30+ other Nature journals & 10,866 & 5.0 & 3.83 & journal \\
& \SetCell[r=3]{l} Other & eLife & 941 & 0.4 & 3.91 & journal \\
& & ArXiv & 3,483 & 1.6 & 3.73 & arxiv \\
& & Other (residual) & 1,075 & 0.5 & 3.52 & mixed \\
\SetCell[r=11]{l} Technical
& & Linux & 7,455 & 3.4 & 3.88 & github \\
& & Web & 6,029 & 2.8 & 3.87 & mixed \\
& & Programming & 4,071 & 1.9 & 4.15 & github \\
& & DevTools & 2,243 & 1.0 & 4.21 & github \\
& & Security & 1,182 & 0.5 & 4.05 & github \\
& & Observability & 731 & 0.3 & 4.07 & github \\
& & Data & 669 & 0.3 & 3.96 & github \\
& & Cloud & 654 & 0.3 & 3.99 & github \\
& & ML & 442 & 0.2 & 3.84 & github \\
& & LLM & 289 & 0.1 & 4.03 & github \\
& & Other Tech & 716 & 0.3 & 3.39 & mixed \\
\SetCell[r=3]{l} Community
& \SetCell[r=2]{l} GitHub SkillMD & Code search & 90,052 & 41.5 & n/a & repo \\
& & Curated lite & 932 & 0.4 & n/a & repo \\
& ClawHub & Community marketplace & 11,389 & 5.2 & n/a & community \\
& & \textbf{Total} & \textbf{216,938} & \textbf{100} & \textbf{3.91} & \\
\end{tblr}
\end{table}

\begin{figure}[t]
\centering
\includegraphics[width=\textwidth]{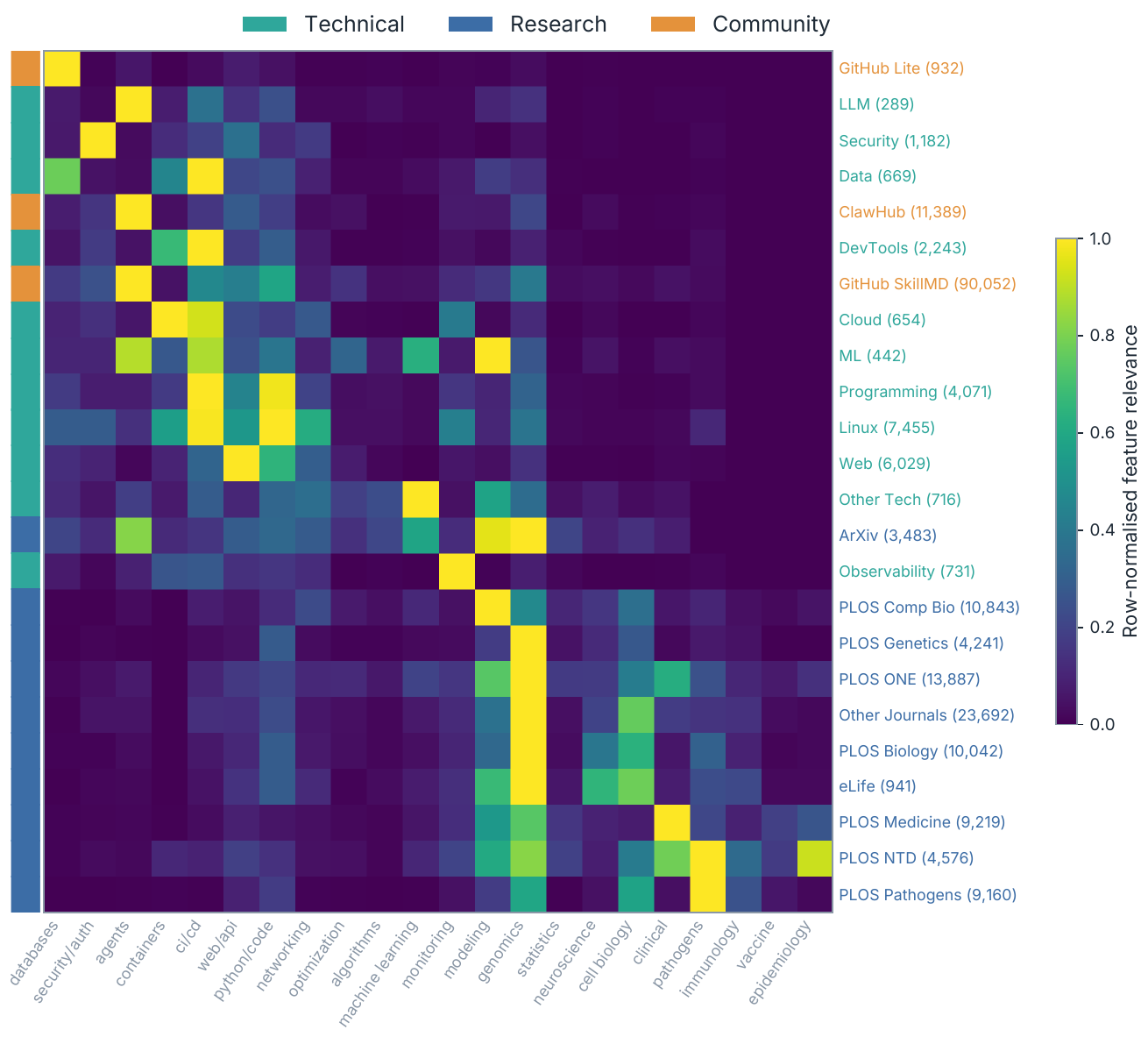}
\caption{Concept-feature profiles of the 24 corpus bundles. Rows (bundles) and columns (22 concept features) are reordered by seriation so that high relevance concentrates along the diagonal; each cell shows the row-normalized relevance (0 to 1) of a concept to a bundle, so every bundle's dominant concepts read brightest. The left strip marks each bundle's macro-category (Technical, Research, or Community; see legend). The resulting block-diagonal structure shows that each bundle activates primarily its own concept group---infrastructure and software concepts for the technical and community bundles (top-left), biomedical and public-health concepts for the research journals (bottom-right)---helping users identify which skill domains are relevant to a given task type.}
\label{fig:taxonomy}
\end{figure}

\paragraph{Bundle Taxonomy and Assignment.}
The 24 bundles follow two organizing principles. Technical, web, and forum skills are grouped by \emph{topic domain}: the pipeline defines 13 domain configurations (e.g., cloud, linux, web; the full list and schema are in Appendix~\ref{appendix:pipeline-params}), each delimiting a domain through host allowlists, GitHub search constraints, and forum tags (\S\ref{sec:source-acq}). Research skills are instead grouped by \emph{publication venue}, namely the Public Library of Science (PLOS), eLife, and the Nature-family journals, which form a natural and citable partition. Each skill inherits the bundle of the configuration that acquired its source, so it belongs to exactly one bundle; topics that do not map unambiguously to a single domain are routed to the most appropriate configuration by an LLM classifier (\S\ref{sec:source-acq}). That this partition is clean in practice is corroborated by the redundancy analysis: cross-bundle near-duplicates are rare (only about 3\% of near-duplicate pairs at Jaccard $\geq 0.8$ span two bundles; \S\ref{sec:dedup}), so the bundles do not substantially overlap. The residual ``other'' categories collect sources that fall within the pipeline's scope but match no specific domain configuration.

We split the corpus into research and technical deliberately. Academic papers provide depth: each paper yields multiple skill kinds covering experimental methodology, theoretical framing, and visualization. Technical sources provide breadth: GitHub repositories, web documentation, and forum discussions capture the operational knowledge agents need for writing, deploying, and maintaining software. A computational biology agent, for instance, needs both the methodological rigor of journal-derived skills and the systems-level guidance of technical skills. Ordering the 24 bundles by their concept profiles recovers this research/technical/community structure (Figure~\ref{fig:taxonomy}).

\subsection{Academic Paper Sources}
\label{sec:academic-sources}

Academic papers make up 41.5\% of the library (90,084 skills), drawn from more than ten peer-reviewed journals and the ArXiv preprint server. Peer-reviewed literature has the highest density of structured, citable knowledge per document, so we prioritized it. Each paper is decomposed into multiple skill kinds covering intellectual framing, experimental design, methodological details, and visual communication. The corpus records what a paper found together with how it framed its question, tested its hypothesis, and presented its results.

\subsubsection*{The PLOS Journal Family}

The PLOS journals collectively contribute 61,968 skills (28.6\% of the library), making them the single largest journal source category. We selected PLOS journals for three reasons: they are fully open-access under Creative Commons licenses, they span a wide range of biological and medical disciplines, and their XML-based publication format enables reliable automated parsing. Seven PLOS journals are represented: PLOS ONE, PLOS Computational Biology, PLOS Biology, PLOS Genetics, PLOS Medicine, PLOS Pathogens, and PLOS Neglected Tropical Diseases (NTD).

For each PLOS paper, the framework extracts four skill kinds that together provide a multi-dimensional decomposition of the paper's content:

\begin{itemize}[leftmargin=2em]
    \item \textbf{paper.idea\_intro} captures the paper's intellectual contribution: the research question, the gap it addresses, and the framing of its hypothesis within the broader literature.
    \item \textbf{paper.experiment} encodes the experimental design, including protocols, controls, sample sizes, and the logic connecting experimental choices to the research question.
    \item \textbf{paper.method} distills the computational or analytical methodology, including algorithms, statistical tests, software tools, and parameter choices.
    \item \textbf{paper.picture} analyzes the paper's figures and visual communication strategy, extracting guidance on how to effectively present scientific data.
\end{itemize}

\begin{figure}[!htbp]
\centering
\includegraphics[width=\textwidth]{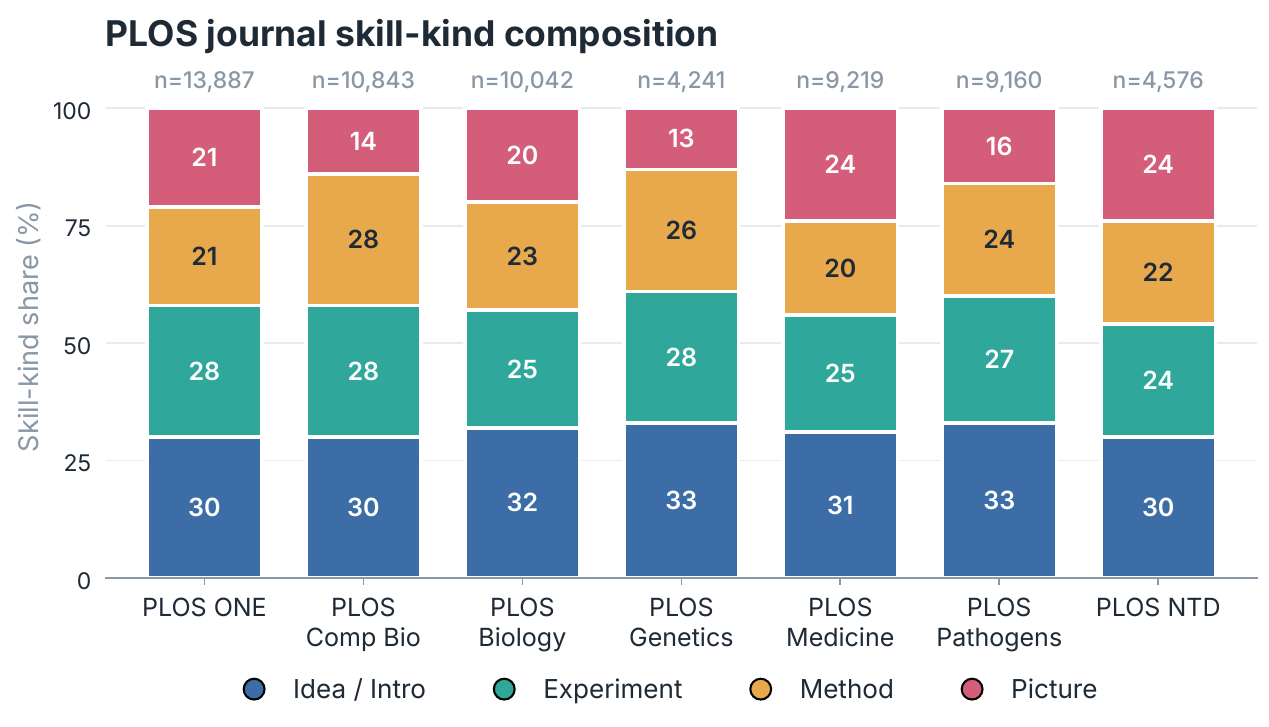}
\caption{Skill kind composition across PLOS journals. Each bar shows the percentage of skills in four categories: idea/intro, experiment, method, and picture. Comp Bio and Genetics tilt toward methods; Medicine and NTD toward visual communication.}
\label{fig:plos-breakdown}
\end{figure}

Table~\ref{tab:plos-breakdown} presents a detailed breakdown of each PLOS journal, including skill counts, average quality scores, and the distribution of skill kinds; Figure~\ref{fig:plos-breakdown} visualizes the skill-kind composition across the seven journals.

\begin{table}[!htbp]
\centering
\caption{PLOS journals detailed breakdown. Percentages indicate per-journal skill kind distribution.}
\label{tab:plos-breakdown}
\small
\begin{tblr}{
  colspec    = {lrrlll},
  row{1}     = {bg=tableHeader, fg=white, font=\bfseries\sffamily},
  row{even}  = {bg=tableStripe},
  rowsep     = 4pt,
  colsep     = 10pt,
  hline{1,Z} = {0.09em, tableRule},
}
Journal & Skills & Score & Idea Intro & Experiment & Method / Picture \\
PLOS ONE & 13,887 & 4.01 & 30\% & 28\% & 21\% / 21\% \\
PLOS Comp Bio & 10,843 & 3.85 & 30\% & 28\% & 28\% / 14\% \\
PLOS Biology & 10,042 & 3.96 & 32\% & 25\% & 23\% / 20\% \\
PLOS Genetics & 4,241 & 3.77 & 33\% & 28\% & 26\% / 13\% \\
PLOS Medicine & 9,219 & 3.94 & 31\% & 25\% & 20\% / 24\% \\
PLOS Pathogens & 9,160 & 3.86 & 33\% & 27\% & 24\% / 16\% \\
PLOS NTD & 4,576 & 3.87 & 30\% & 24\% & 22\% / 24\% \\
\end{tblr}
\end{table}

\textbf{PLOS ONE} is the largest single journal bundle (13,887 skills, avg 4.01) and the only multidisciplinary PLOS journal, spanning biology, medicine, engineering, and social sciences. Its skill kind distribution is nearly uniform (30/28/21/21\%), consistent with its broad disciplinary scope. \textbf{PLOS Biology} (10,042 skills, avg 3.96) is similar (32/25/23/20\%), a selective journal that publishes conceptual and experimental work in roughly equal measure.

The other five journals are more specialized, and the skill kind distributions show it. \textbf{PLOS Computational Biology} (10,843 skills, avg 3.85) tilts toward methods (28\%) and away from figures (14\%), consistent with its computational and mathematical emphasis. \textbf{PLOS Genetics} (4,241 skills, avg 3.77) follows a similar pattern: method-heavy (26\%) with low picture share (13\%), consistent with a computational and statistical emphasis. \textbf{PLOS Medicine} (9,219 skills, avg 3.94) goes the opposite direction: 24\% picture, the highest in the family. Clinical papers rely heavily on Kaplan-Meier curves and forest plots. \textbf{PLOS Pathogens} (9,160 skills, avg 3.86) leads on idea\_intro (33\%). Papers in this journal spend considerable space framing novel hypotheses about host-pathogen interactions. \textbf{PLOS NTD} (4,576 skills, avg 3.87), the smallest bundle, resembles PLOS Medicine in its high picture share (24\%): epidemiological maps and geographic visualizations are central to the field.

\subsubsection*{Nature Family and Other Journals}

The Nature family of journals contributes 22,617 skills (10.4\% of the library), making it the second largest journal source category after PLOS. We decomposed the formerly monolithic ``Other Journals'' bundle into its constituent sub-journals using DOI-based classification against the source URLs. The result reveals over 30 distinct Nature-branded journals, with the top eleven each contributing over 900 skills: Nat.\ Neuroscience (1,303), Nat.\ Cell Biology (1,163), Nat.\ Methods (1,160), Nat.\ Immunology (1,084), Nat.\ Microbiology (1,073), Nature flagship (1,048), Scientific Reports (1,030), Nat.\ Medicine (998), Nat.\ Human Behaviour (997), Nat.\ Chemistry (952), and Nat.\ Metabolism (943). The remaining 30+ journals collectively contribute 10,866 skills, including titles such as Nat.\ Structural \& Molecular Biology, Nat.\ Plants, and Nat.\ Ecology \& Evolution.

Quality scores run high across the Nature family. The Nature flagship journal achieves the highest average among all research sources (4.12), reflecting the depth and precision of its papers. Nat.\ Methods (4.05) ranks second, consistent with its focus on detailed methodological descriptions that map naturally to actionable skills. The family-wide average of 3.83 is competitive with PLOS despite the higher selectivity of Nature journals producing fewer but more focused skills.

The idea\_intro share across Nature journals (36.1\%) is the highest of any source family, consistent with these journals' emphasis on conceptually novel work.

\textbf{eLife} (941 skills, avg 3.91) is selective and prizes conceptual novelty. idea\_intro accounts for 34.9\% of its skills, while picture is only 10.3\%. A residual category of 1,075 skills (0.5\%) includes PLOS cover image descriptions and a small number of items that did not match any journal family pattern.

\subsubsection*{ArXiv Preprints}

ArXiv contributes 3,483 skills (1.6\%, avg 3.73, the lowest among research sources, as expected for non-peer-reviewed preprints). Despite the lower scores, ArXiv fills an important niche by providing access to recent methods before journal publication. The skill kind taxonomy for ArXiv differs from the journal taxonomy; Table~\ref{tab:kind-comparison} compares the two systems.

\begin{table}[!htbp]
\centering
\caption{Comparison of skill kinds between journal and ArXiv sources.}
\label{tab:kind-comparison}
\small
\begin{tblr}{
  colspec    = {lll},
  row{1}     = {bg=tableHeader, fg=white, font=\bfseries\sffamily},
  row{even}  = {bg=tableStripe},
  row{6}     = {bg=codeBg},
  rowsep     = 4pt,
  colsep     = 14pt,
  hline{1,Z} = {0.09em, tableRule},
  hline{6}   = {0.05em},
}
Dimension & Journal Papers & ArXiv Preprints \\
Idea / Framing & paper.idea\_intro (31\%) & paper\_writing (34\%) \\
Methodology & paper.method (22\%) & paper\_writeup (32\%) \\
Experiments & paper.experiment (26\%) & experiment\_design (31\%) \\
Visualization & paper.picture (21\%) & results\_analysis (2.4\%) \\
Focus & Understanding \& introducing & Writing \& reproducing \\
\end{tblr}
\end{table}

The two taxonomies reflect different use cases. Journal skill kinds center on understanding a published paper: paper.idea\_intro captures the research question, paper.method distills the methodology. ArXiv skill kinds are oriented toward writing and reproducing research: paper\_writing captures how to frame a contribution, paper\_writeup distills methods into reproducible recipes, and experiment\_design focuses on experimental setup. The small share of results\_analysis (2.4\%) reflects the predominantly methodological nature of preprints.

\subsection{Technical Sources}
\label{sec:technical-sources}

Technical sources account for 11.3\% of the library (24,481 skills), drawn from GitHub repositories, web pages, Stack Overflow forums, and ArXiv papers in technical domains. Unlike academic sources with consistent structure, technical sources vary widely in format, depth, and quality, presenting both opportunities for capturing hands-on operational knowledge and challenges from variable signal-to-noise ratios.

\subsubsection*{GitHub Repositories}

GitHub repositories constitute the largest technical source, contributing 20,064 skills (9.2\% of the library).\footnote{Seven further GitHub-sourced skills fall in research bundles; the by-source-type totals in Table~\ref{tab:pipeline-funnel} therefore list 20,071 GitHub skills across the full pipeline subset.} The collection pipeline uses domain-configured search queries with minimum star count and pushed\_after date thresholds to identify actively maintained, community-validated repositories. For each qualifying repository, the framework extracts the README file and key code files, then generates skills that capture the repository's core functionality, setup procedures, and usage patterns.

Table~\ref{tab:github-domains} presents the distribution of GitHub-derived skills across technical domains.

\begin{table}[!htbp]
\centering
\caption{GitHub skills by domain across the full pipeline GitHub subset (20,071 skills). The 20,064 figure in the text counts only skills published in technical bundles, excluding seven GitHub-sourced skills routed to research bundles. ``Top Contributors'' lists the most frequently occurring repository owners.}
\label{tab:github-domains}
\small
\begin{tblr}{
  colspec    = {lrrl},
  row{1}     = {bg=tableHeader, fg=white, font=\bfseries\sffamily},
  row{even}  = {bg=tableStripe},
  rowsep     = 4pt,
  colsep     = 14pt,
  hline{1,Z} = {0.09em, tableRule},
}
Domain & Skills & Avg Score & Top Contributors \\
linux & 7,262 & 3.90 & Individual developers (various) \\
programming & 3,874 & 4.15 & rust-lang, golang, python \\
web & 2,810 & 3.93 & vercel, facebook, vuejs \\
devtools & 2,001 & 4.23 & microsoft, JetBrains, docker \\
security & 1,127 & 4.08 & OWASP, zaproxy, snort \\
observability & 680 & 4.09 & grafana, prometheus, elastic \\
data & 600 & 3.98 & apache, databricks \\
cloud & 532 & 4.03 & kubernetes, aws, hashicorp \\
ml & 365 & 3.85 & huggingface, pytorch, tensorflow \\
llm & 250 & 4.06 & langchain, openai, llamaindex \\
other & 570 & 3.25 & various \\
\end{tblr}
\end{table}

\textbf{DevTools} has the highest average score (4.23) despite modest volume. Microsoft, JetBrains, and Docker write thorough READMEs, which in turn yield good skills. \textbf{Linux} is the largest domain (7,262 skills) but scores lower (3.90): individual maintainers vary widely in how much they document. \textbf{Programming} (avg 4.15) benefits from language ecosystem repositories (rust-lang, golang, python) with unusually complete documentation.

Community-driven and enterprise-driven domains differ, but not cleanly by ownership. Enterprise repositories cluster at the high end (DevTools 4.23, Observability 4.09, Cloud 4.03), probably because corporate documentation teams write more structured READMEs; but the community-driven programming ecosystem is comparably high (4.15), so this is a tendency, not a rule.

\subsubsection*{Web Pages}

Web pages contribute 3,130 skills (1.4\% of the library), with approximately 95\% concentrated in the web development domain. Table~\ref{tab:web-sources} lists the top 10 source domains by skill count.

\begin{table}[!htbp]
\centering
\caption{Top 10 web page source domains by skill count.}
\label{tab:web-sources}
\small
\begin{tblr}{
  colspec    = {lrr},
  row{1}     = {bg=tableHeader, fg=white, font=\bfseries\sffamily},
  row{even}  = {bg=tableStripe},
  rowsep     = 3pt,
  colsep     = 14pt,
  hline{1,Z} = {0.09em, tableRule},
}
Source Domain & Skills & Avg Score \\
developer.mozilla.org (MDN) & 145 & 3.78 \\
dev.to & 137 & 3.86 \\
github.com & 117 & 3.66 \\
blog.logrocket.com & 62 & 3.44 \\
geeksforgeeks.org & 56 & 3.96 \\
nextjs.org & 53 & 3.55 \\
freecodecamp.org & 41 & 3.61 \\
oneuptime.com & 40 & 3.40 \\
web.dev & 36 & 3.42 \\
linkedin.com & 34 & 3.29 \\
\end{tblr}
\end{table}

The distribution has a long tail. The 3,130 web page skills come from 1,326 unique source domains; the top 10 account for 721 skills, while the remaining 2,409 originate from 1,316 domains at an average of 1.8 skills each, and 976 domains contribute a single skill. Of the 3,130 skills, 2,980 (95.2\%) fall in the web-development domain, with the remainder scattered across the Linux, DevTools, and other technical domains.

The overall web page average is 3.76, lower than GitHub's 4.01. For non-web domains, web page skills concentrate on official documentation sites, and the average score for cloud-domain web pages drops to 2.25 (across only eight skills), consistent with diminishing returns from web scraping in domains with sparse tutorial coverage.

\subsubsection*{Stack Overflow Forums}

Stack Overflow contributes 1,162 skills (0.5\% of the library), all sourced from stackoverflow.com. Stack Overflow content is published under CC-BY-SA licenses (versions 3.0 and 4.0, depending on the post date), which require attribution and share-alike compliance for redistribution; the published bundle retains the source URL and attribution metadata needed to support these obligations. Table~\ref{tab:forum-domains} shows the domain breakdown.

\begin{table}[!htbp]
\centering
\caption{Stack Overflow forum skills by domain.}
\label{tab:forum-domains}
\small
\begin{tblr}{
  colspec    = {lrr},
  row{1}     = {bg=tableHeader, fg=white, font=\bfseries\sffamily},
  row{even}  = {bg=tableStripe},
  rowsep     = 4pt,
  colsep     = 18pt,
  hline{1,Z} = {0.09em, tableRule},
}
Domain & Skills & Avg Score \\
web & 941 & 3.83 \\
other & 168 & 3.12 \\
linux & 52 & 3.42 \\
cloud & 1 & 3.00 \\
\end{tblr}
\end{table}

The web domain dominates Stack Overflow-derived skills at 81\%, reflecting the platform's historically strong coverage of frontend and web development topics. The Q\&A format maps naturally to the skill structure: a question provides the background and problem context, the accepted answer supplies the solution steps, and comments add verification criteria or edge-case warnings.

However, the moderate average scores (3.12 to 3.83) reveal a limitation of forum-derived skills. Stack Overflow answers are often terse, version-specific, or narrowly scoped to the original question's context, limiting their generalizability. The ``other'' domain scores lowest (3.12), as miscellaneous topics tend to produce answers too niche for effective skill extraction. Despite these limitations, Stack Overflow's CC-BY-SA licensing (with attribution and share-alike obligations) and problem-and-solution structure make it a valuable complement to GitHub and web page sources.

\subsubsection*{ArXiv Preprints in Technical Domains}

\textbf{ArXiv preprints in technical domains} contribute 125 skills, split between the Linux domain (64 skills, avg 2.80) and the ML domain (61 skills, avg 3.43). These two domains behave very differently: academic papers about operating system concepts and kernel design do not readily convert to actionable, step-by-step skills, while ML methodology papers describing model architectures and training procedures are inherently procedural and convert naturally. The linux-domain ArXiv average of 2.80 is among the lowest source-level averages in the corpus, ahead of only cloud-domain web pages (2.25).

\subsection{Community and Repository Bundles}
\label{sec:community}

Beyond the SkillGate pipeline, the released library integrates two large, pre-existing skill collections that are harvested rather than generated. Together they contribute 102,373 skills (47.2\% of the library) and are reported as the \emph{Community} category in Table~\ref{tab:corpus-summary}.

\textbf{GitHub SkillMD} (90,984 skills, source type \texttt{repo}) is a bulk harvest of \texttt{SKILL.md} files already authored and committed to public GitHub repositories, located via code search rather than produced by our generation templates. Because these skills are human-authored artifacts that ship inside agent projects, they are admitted directly; they carry a uniform placeholder quality marker rather than a GPT-assigned score on the 1-to-5 scale, and are therefore excluded from the score analyses in \S\ref{sec:quality-scores} and Table~\ref{tab:score-distribution}. This bundle is the single largest in the library and substantially broadens coverage of real-world, repository-embedded agent skills.

\textbf{ClawHub community} (11,389 skills, source type \texttt{claw-*}) mirrors the ClawHub marketplace~\cite{clawhub2026} of community-contributed skills. ClawHub items are scored on a separate marketplace scale (not the 1-to-5 publish-gate rubric) and are likewise excluded from the pipeline score statistics. Integrating the marketplace makes the library a superset of the community ecosystem while keeping provenance explicit per bundle.

Both integrated collections bypass SkillGate, generation, and the publish gate; the framework described in \S\ref{sec:framework} applies only to the 114,565 pipeline-produced Research and Technical skills. We keep the two populations clearly separated throughout so that downstream users can opt into the source-grounded pipeline subset, the community subset, or the full library.

\subsection{Deduplication and Redundancy}
\label{sec:dedup}

A library of this size invites an obvious question: are the 216,938 skills genuinely distinct, or does the headline count conceal duplicates? We measure redundancy at three levels (Table~\ref{tab:dedup}), recomputed directly from the released SQLite bundles. First, \textbf{identity}: \texttt{skill\_id} is a global primary key, all 216,938 published skills carry distinct IDs, and the 24 bundles are pairwise disjoint, so no skill is counted twice and none appears in two bundles. Second, \textbf{verbatim content}: 9,831 skills (4.5\%) share byte-identical normalized text with at least one other skill, that is, distinct IDs that nonetheless render to the same body, such as shared templates or repeated boilerplate. Third, \textbf{near-duplication}: we compute a 128-permutation MinHash over word 5-shingles of each skill body and retrieve near-duplicate pairs with banded locality-sensitive hashing (LSH), then cluster skills whose estimated Jaccard similarity exceeds a threshold. At Jaccard $\geq 0.8$, 24,988 skills (11.5\%) lie in a near-duplicate cluster.

\begin{table}[!htbp]
\centering
\caption{Redundancy in the SkillCenter corpus at three levels: identical IDs, verbatim content, and lexical near-duplication (128-permutation MinHash over word 5-shingles with LSH retrieval, Jaccard $\geq 0.8$). Shares are of the full library (216,938 skills).}
\label{tab:dedup}
\small
\begin{tblr}{
  colspec    = {llrr},
  row{1}     = {bg=tableHeader, fg=white, font=\bfseries\sffamily},
  row{even}  = {bg=tableStripe},
  rowsep     = 4pt,
  colsep     = 16pt,
  hline{1,Z} = {0.09em, tableRule},
}
Redundancy level & Method & Skills & Share \\
Duplicate skill IDs & primary-key match & 0 & 0.00\% \\
Verbatim content & normalized hash & 9,831 & 4.53\% \\
Near-duplicate (Jaccard $\geq$ 0.8) & MinHash + LSH & 24,988 & 11.52\% \\
\end{tblr}
\end{table}

Redundancy is highly uneven across the library and is confined almost entirely to the externally harvested community bundles (Table~\ref{tab:dedup-by-domain}). The SkillGate pipeline subset is essentially duplicate-free: every one of its 21 domains sits at or below 0.01\%, reflecting per-source generation and source-grounding. The GitHub SkillMD bundle, by contrast, is about 27\% near-duplicate (and its small curated-lite subset reaches 65\%), as expected for \texttt{SKILL.md} files mined from public repositories, which include forks, templates, and copied boilerplate; the ClawHub marketplace sits at about 5\%. Near-duplicate pairs are overwhelmingly \emph{within} a single bundle: only about 3\% of near-duplicate pairs at Jaccard $\geq 0.8$ span two different bundles, confirming that the 24 bundles do not overlap one another and that the corpus total is not inflated by cross-bundle repetition. Table~\ref{tab:dedup-examples} gives representative duplicate groups, spanning verbatim copies and small-edit near-duplicates; Appendix~\ref{appendix:redundancy-audit} lists every domain that contains duplicates, with counts and a concrete example from each.

\begin{table}[!htbp]
\centering
\caption{Near-duplicate concentration by domain (MinHash Jaccard $\geq 0.8$). Redundancy is confined to the externally harvested community bundles; every individual SkillGate pipeline domain (10 research, 11 technical) sits at or below 0.01\%, so they are grouped here. The 932-skill curated-lite GitHub subset is the most redundant, consistent with hand-picked popular skills.}
\label{tab:dedup-by-domain}
\small
\begin{tblr}{
  colspec    = {lrr},
  row{1}     = {bg=tableHeader, fg=white, font=\bfseries\sffamily},
  row{even}  = {bg=tableStripe},
  rowsep     = 3pt,
  colsep     = 16pt,
  hline{1,Z} = {0.09em, tableRule},
}
Domain group & Skills & Near-dup \% \\
Pipeline Research (10 domains) & 90,084 & $<$0.01 \\
Pipeline Technical (11 domains) & 24,481 & 0.00 \\
GitHub SkillMD (code search) & 90,052 & 26.9 \\
GitHub SkillMD (curated lite) & 932 & 65.0 \\
ClawHub marketplace & 11,389 & 5.4 \\
\end{tblr}
\end{table}

\begin{table}[!htbp]
\centering
\caption{Representative duplicate skills from the GitHub SkillMD community bundle, where almost all redundancy concentrates. Verbatim groups share byte-identical bodies; near-duplicate similarity is the MinHash-estimated Jaccard.}
\label{tab:dedup-examples}
\small
\begin{tblr}{
  colspec    = {llp{5.0cm}p{4.6cm}},
  row{1}     = {bg=tableHeader, fg=white, font=\bfseries\sffamily},
  row{even}  = {bg=tableStripe},
  column{3,4} = {halign=l},
  rowsep     = 3pt,
  colsep     = 8pt,
  hline{1,Z} = {0.09em, tableRule},
}
Type & Sim. & Representative example & Why it repeats \\
Verbatim & identical & ``PPTX creation, editing, and analysis'' found in 41 repositories & A popular skill forked into many skill-collection repos \\
Verbatim & identical & Stub whose entire body is ``Follow the instructions in ./workflow.md'' & Framework-generated placeholder stubs \\
Verbatim & identical & Unmodified ``Insert overview text here'' default rule & Uncommitted scaffold boilerplate \\
Near-dup & $\approx$0.90 & Two ``eddication'' skills, one extending the other's capability list & A fork that appends extra content \\
Near-dup & $\approx$0.83 & Two ``fast-io'' skills differing only in version string and date & A version bump of the same skill \\
\end{tblr}
\end{table}

Two caveats apply. First, MinHash with Jaccard similarity captures \emph{lexical} near-duplication; meaning-preserving paraphrases are not detected, so the figures above are a lower bound on true redundancy.

Second, we also tested embedding-based \emph{semantic} deduplication~\cite{abbas2023semdedup} to catch paraphrases that MinHash misses, but found it unreliable on this corpus: across seven off-the-shelf sentence encoders~\cite{reimers2019sentencebert} spanning 22M to 8B parameters (from the MiniLM, BGE, GTE, E5, and Qwen3-Embedding families), the share of skills flagged as near-duplicate at a fixed cosine-0.9 threshold ranged all the way from 49\% to 93\%, reflecting each encoder's embedding geometry rather than any property of the data; manual inspection confirmed that the high-similarity pairs are topically related but genuinely distinct skills, not paraphrases. We therefore report the lexical MinHash figures of Table~\ref{tab:dedup} as our defensible redundancy measure and leave a calibrated semantic estimate to future work. The residual redundancy concentrates in the harvested community bundle, and users who require a deduplicated view can restrict to the pipeline subset or apply standard near-duplicate filtering.

\subsection{Quality Score Analysis}
\label{sec:quality-scores}

Each skill in the pipeline subset receives an integer quality score from 1 to 5, assigned by GPT-5.2 during the publish gate; the integrated community collections use separate scales and are excluded here. The rubric evaluates clarity, accuracy, actionability, and evidence grounding. The average over these 114,565 scored skills is 3.91, but the distribution is heavily skewed: approximately 82.1\% of them receive a score of exactly 4. Score 3 and score 5 together account for most of the remainder. The model rarely commits to an extreme rating. This score inflation is consistent with documented biases in LLM-as-judge evaluation, where models exhibit self-enhancement and verbosity effects that compress scores toward the high end~\cite{zheng2023judge, liu2023geval}.

Domain-level averages occupy a narrow band: most technical domains fall within roughly 4.0 to 4.2 (e.g., DevTools 4.21, Programming 4.15, LLM 4.03), a spread small enough that, given the single-LLM scorer and the heavy score-4 collapse, we do not read the ordering among them as meaningful. The only clearly separated case is the residual ``other'' domain (3.39), a catch-all for miscellaneous content.

Source-type differences are likewise subtle. The high-quality source types cluster closely (GitHub 4.01, journals 3.90, ArXiv 3.71), too near to rank reliably. The full per-source, per-score breakdown is deferred to Table~\ref{tab:score-distribution} in \S\ref{sec:publishing}; we report these averages for completeness rather than as evidence of relative quality, and whether the finer differences hold under human evaluation is an open question (\S\ref{sec:quality-enhancement}).

The publish gate is configured to exclude skills scoring below 3.0, removing approximately 3\% of generated skills; a small number of early-run skills predate full enforcement of this threshold (see \S\ref{sec:publishing}). The threshold is conservative: it filters the clearly defective tail while retaining the broad score-4 middle that represents skills of adequate quality. Section~\ref{sec:future-work} discusses calibration strategies including human annotation campaigns and multi-model consensus scoring.

\subsection{Search System and Agent Integration}
\label{sec:search-integration}

The distribution system obeys three constraints. First, it uses no embedding models, no vector databases, and no network access at query time, so it has zero external dependencies. Second, agents install only the domains they need, which keeps the system modular at the domain level. Third, skill retrieval requires no explicit user action, so the integration is invisible to the user.

\subsubsection*{Architecture}

The search system is built on SQLite FTS5~\cite{sqlite_fts5}, a full-text search extension that ships with SQLite and requires no external dependencies. Each domain bundle is a self-contained SQLite database file, with technical domain bundles averaging approximately 120\,MB and research domain bundles averaging approximately 1\,GB. In the current release the FTS5 index covers each skill's title and domain rather than its full body; queries therefore match against the skill's declared topic, and matching against full skill content is a planned extension (\S\ref{sec:roadmap}).

Each bundle uses a four-table schema (detailed in Appendix~\ref{appendix:reproducibility}) that separates metadata from content, enabling lazy loading: the search engine scans lightweight index entries and fetches full skill markdown only for selected results.

We chose FTS5 over vector search for four reasons. First, FTS5 is fully offline: it requires no embedding model download, no GPU, and no network access, making it suitable for air-gapped environments. Second, agent queries tend to be short keyword phrases (``configure nginx reverse proxy,'' ``pandas groupby aggregation'') that BM25, the standard bag-of-words ranking function used in full-text search, matches against each skill's title and domain. Third, FTS5 ranking is deterministic and auditable, a key property for debugging agent behavior. Fourth, storage overhead is minimal compared to dense vector indexes. The main cost of this design is retrieval recall: because only the title and domain are indexed, a query whose terms appear in a skill's body but not its title will miss it, which our downstream study (\S\ref{sec:downstream}) identifies as the binding constraint in practice.

\subsubsection*{Query Pipeline}

A \texttt{skill-search} command accepts a natural language query string and runs it through four stages:
\begin{enumerate}[leftmargin=2em]
    \item The FTS5 engine returns candidate skills ranked by BM25 score.
    \item Optional filters narrow results by domain, skill kind, or minimum quality score.
    \item When multiple bundles are installed, the pipeline searches all of them in parallel, merge-sorts by BM25, and deduplicates.
    \item Content loads on demand: the initial query returns only metadata from \texttt{skills\_index}, and full skill markdown is fetched from \texttt{skills\_content} only when the agent picks a specific skill.
\end{enumerate}
A typical search scans thousands of index entries but loads only two or three full documents.

\subsubsection*{Agent Integration Design}

The agent integration layer automates three steps: project type detection (by inspecting configuration files such as \texttt{package.json} or \texttt{pyproject.toml}), bundle installation from Hugging Face~\cite{huggingface_datasets}, and skill injection into the LLM context.

During task execution, the agent follows a retrieve-then-act pattern: it issues a skill search based on the task description, selects the highest-scoring relevant skills, and injects their content into the LLM context. Source URLs are stripped from the injected prompt to prevent agents from generating stale or hallucinated links; the published bundle nonetheless retains each skill's source URL and attribution metadata, so license obligations such as CC-BY-SA attribution can still be satisfied. The entire integration is invisible to the end user: the agent retrieves and applies skills without manual configuration. Because retrieval is a local SQLite FTS5 query over lightweight index entries, it adds negligible latency to the agent loop; a systematic benchmark across bundle sizes and multi-bundle configurations is planned (\S\ref{sec:roadmap}).

\section{The SkillCenter Collection Framework}
\label{sec:framework}

The framework converts raw source material into searchable, quality-checked skills through five stages.

\subsection{Framework Overview}

\begin{figure*}[!htbp]
\centering
\includegraphics[width=\textwidth]{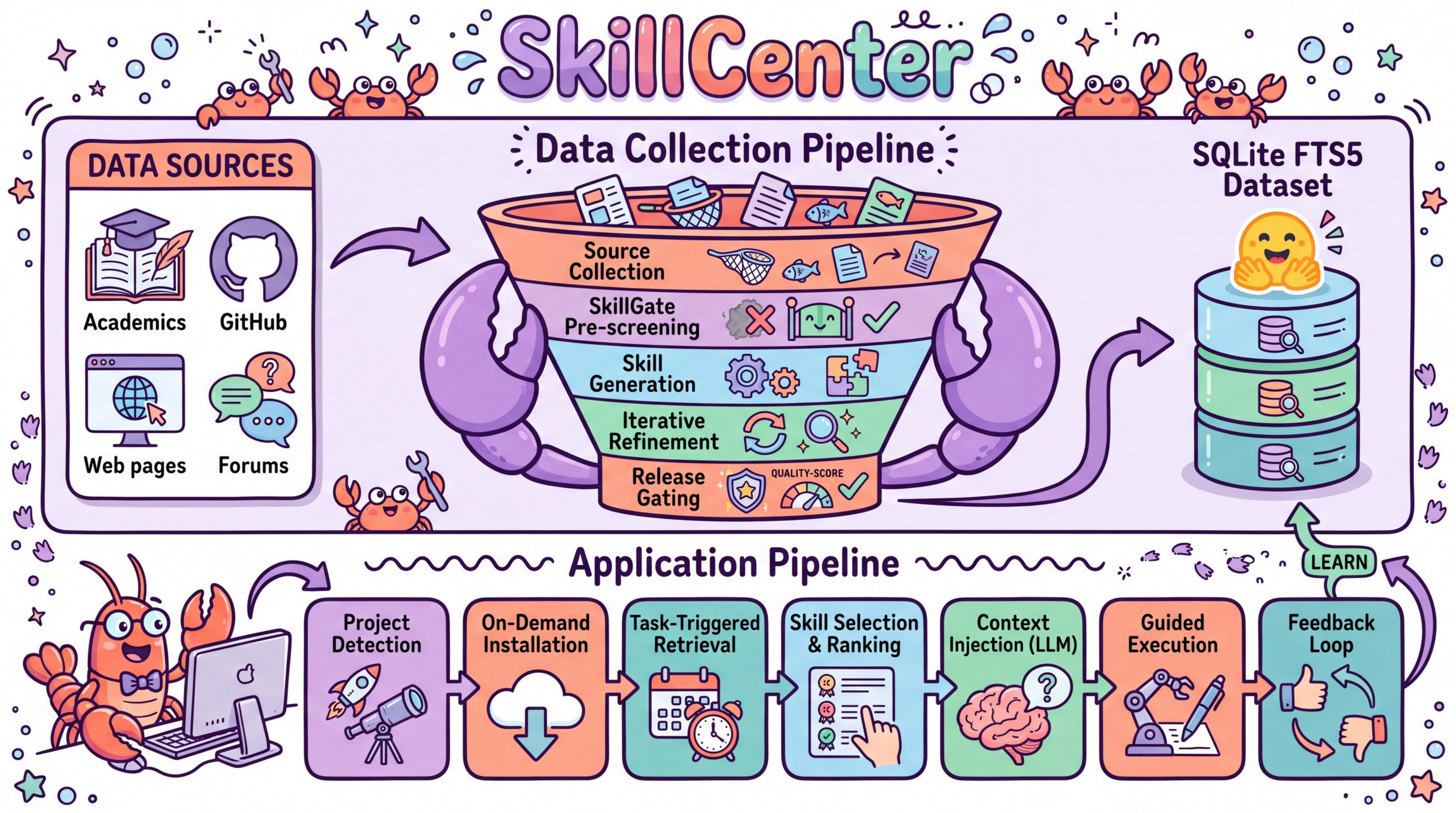}
\caption{Overview of the SkillCenter pipeline. The data collection pipeline (top) ingests sources from academic publications, GitHub, web pages, and forums, then applies SkillGate filtering, template-driven generation, iterative refinement, and quality-controlled release gating to produce SQLite FTS5 skill bundles distributed via Hugging Face. The application pipeline (bottom) consumes these skills at inference time through project detection, on-demand installation, BM25-based retrieval, skill ranking, LLM context injection, guided execution, and a feedback loop.}
\label{fig:pipeline}
\end{figure*}

The framework is organized as a five-stage pipeline (Figure~\ref{fig:pipeline}).

The five stages are:

\begin{enumerate}[leftmargin=2em]
    \item \textbf{Source Acquisition} (\S\ref{sec:source-acq}): Multi-source fetching with domain-aware routing. The framework retrieves content from web pages, GitHub repositories, Stack Overflow forums, ArXiv preprints, and journal APIs, applying per-source rate limiting and text extraction to produce uniform capture artifacts.
    \item \textbf{Quality Gate} (\S\ref{sec:quality-gen}): An LLM-based pre-filter called SkillGate evaluates whether each source document is suitable for skill extraction before any generation tokens are spent. Sources are classified as pass, maybe, or fail based on actionability, reproducibility, and content density.
    \item \textbf{Skill Generation} (\S\ref{sec:quality-gen}): Template-driven LLM generation produces structured skill documents using kind-specific prompts. Four template types tailor the generation to the source material's nature (Table~\ref{tab:source-routing}).
    \item \textbf{Iterative Improvement} (\S\ref{sec:iterative-improvement}): Each generated skill undergoes up to three refinement passes (configurable). Each pass lints the skill for structural issues, identifies unaddressed improvement suggestions, and rewrites the skill with deterministic source-grounding checks to confirm that claims trace back to the original source.
    \item \textbf{Publishing} (\S\ref{sec:publishing}): A four-criterion publish gate filters on quality score, license compatibility, plagiarism ratio, and placeholder density. Skills that pass are packaged into SQLite FTS5 bundles and distributed via Hugging Face.
\end{enumerate}

Appendix~\ref{appendix:worked-example} traces a single skill (a GitHub README on automated security review) through all five stages, from raw source to published bundle entry.

\subsection{Source Acquisition}
\label{sec:source-acq}

The first stage retrieves raw content from five primary source types. All sources undergo common post-processing: text extraction, truncation to 12,000 characters, URL canonicalization (removing tracking parameters), and persistence as capture artifacts.

\subsubsection*{Input Sources and Fetching}

Table~\ref{tab:source-methods} summarizes the five primary source types and their acquisition methods.

\begin{table}[!htbp]
\centering
\caption{Source acquisition methods. Each source type uses a distinct API and extraction pipeline, unified by a common 12,000-character truncation and URL canonicalization step.}
\label{tab:source-methods}
\small
\begin{tblr}{
  colspec    = {p{2.2cm}p{2.8cm}p{3.2cm}p{3.2cm}p{2.2cm}},
  cells      = {halign=l},
  row{1}     = {bg=tableHeader, fg=white, font=\bfseries\sffamily},
  row{even}  = {bg=tableStripe},
  rowsep     = 4pt,
  colsep     = 4pt,
  hline{1,Z} = {0.09em, tableRule},
}
Source Type & API / Method & Input & Extraction & Rate Control \\
Web Pages & HTTP GET + readability & Seed URLs from domain config & HTML$\to$text, truncate 12{,}000 chars & Per-host delay \\
GitHub & GitHub REST API & Domain queries, min\_stars & README + key code files & API rate limit \\
Stack Overflow & SO API v2.3 & Tagged questions per domain & Q + accepted answer + comments & API quota \\
ArXiv & ArXiv API + PDF & Category-based queries & PDF$\to$text extraction & 3\,s delay \\
Journals & PLOS/eLife XML API & Journal-specific queries & XML structured parsing & Journal-specific \\
\end{tblr}
\end{table}

Each source type follows its own acquisition path. \textbf{Web pages}: HTTP GET with readability extraction and per-host delays. \textbf{GitHub}: domain-configured search queries with minimum star count and pushed\_after filters, extracting README and key code files. \textbf{Stack Overflow}: the Stack Exchange API v2.3, concatenating question, accepted answer, and top comments. \textbf{ArXiv}: the ArXiv API with PDF-to-text conversion. \textbf{Journals} (PLOS, eLife): XML APIs that enable precise section-level extraction.

\subsubsection*{Domain-Aware Configuration}

The framework supports 13 configured domains, each specifying host allowlists, seed URLs, GitHub search constraints, and forum tags (full configuration schema in Appendix~\ref{appendix:pipeline-params}). When a topic does not map unambiguously to a single domain, an LLM-driven classifier routes it to the most appropriate configuration.

\paragraph{Case Study: ``kubernetes networking'' acquisition path.}
The topic ``kubernetes networking'' could belong to the cloud, linux, or web domains. The LLM classifier assigns it to cloud, which triggers three parallel fetches using the cloud domain configuration: web pages from kubernetes.io, GitHub repositories matching ``kubernetes networking'' with min\_stars=100 (e.g., cilium, flannel), and Stack Overflow questions tagged kubernetes+networking. All source documents are truncated to 12,000 characters and saved as capture artifacts.

\subsection{Quality Gate and Skill Generation}
\label{sec:quality-gen}

This subsection describes the two-phase process: a pre-generation quality gate that filters unsuitable sources, and template-driven generation of structured skill documents.

\subsubsection*{SkillGate: Pre-Generation Filter}

SkillGate is an LLM-based pre-filter that evaluates source suitability before spending generation tokens. The gate receives a source excerpt (truncated to 4,000 characters) and produces a three-class verdict: pass, maybe (configurable, default: pass), or fail. Table~\ref{tab:skillgate-criteria} summarizes the criteria, and Appendix~\ref{appendix:prompts} reproduces the gate prompt together with the 1-to-5 quality-score rubric.

\begin{table}[!htbp]
\centering
\caption{SkillGate judgment criteria. The gate classifies each source into one of three verdicts based on actionability and reproducibility.}
\label{tab:skillgate-criteria}
\small
\begin{tblr}{
  colspec    = {p{1.5cm}p{6.5cm}p{5.5cm}},
  cells      = {halign=l},
  row{1}     = {bg=tableHeader, fg=white, font=\bfseries\sffamily},
  row{even}  = {bg=tableStripe},
  rowsep     = 4pt,
  colsep     = 8pt,
  hline{1,Z} = {0.09em, tableRule},
}
Verdict & Criteria & Action \\
pass & Actionable, contains steps/code, reproducible & Proceed to generation \\
maybe & Valuable content but lacks clarity, steps, or verification & Configurable (default: pass) \\
fail & Too short ($<$200 chars), spam/marketing, pure opinion, not actionable & Skip generation \\
\end{tblr}
\end{table}

Sources below 200 characters receive an automatic fail. Above this threshold, the LLM returns a structured output: verdict, a SkillGate suitability score (0 to 10, not the 1-to-5 quality score used at the publish gate), and lists of supporting and opposing signals.

Two contrasting cases show what the gate does in practice.

\paragraph{Case Study A (pass $\to$ score\,=\,5 skill).} The GitHub repository ca-risken/security-review (MIT), containing a complete README with GitHub Actions configuration and input/output descriptions, received a SkillGate verdict of pass (score=8) and produced a score-5 skill titled ``Add RISKEN Security Code Review to GitHub Pull Requests.'' The resulting skill featured a clear Background, four concrete Use Cases, eight numbered Steps with YAML code blocks, and an explicit Verification section.

\paragraph{Case Study B (would-fail $\to$ score\,=\,2 skill).} By contrast, a bare code repository containing only a Mellor-Crummey-Scott (MCS) lock implementation (no README, no context, no usage instructions) would have received a fail verdict had the gate been enabled. The resulting score-2 skill suffered from vague Background text, ``not provided'' placeholders in Inputs, and Steps that merely restated code lines without operational guidance.

Structured source documentation maps directly to skill components. Bare code, without a README or usage instructions, does not.

\subsubsection*{Skill Generation Templates}

Four template types tailor the generation to different source materials (Table~\ref{tab:gen-templates}).

\begin{table}[!htbp]
\centering
\caption{Skill generation templates. Each template defines a system role, required sections, and the source types it applies to.}
\label{tab:gen-templates}
\small
\begin{tblr}{
  colspec    = {p{2.8cm}p{3.2cm}p{5.5cm}p{2.8cm}},
  cells      = {halign=l},
  row{1}     = {bg=tableHeader, fg=white, font=\bfseries\sffamily},
  row{even}  = {bg=tableStripe},
  rowsep     = 4pt,
  colsep     = 4pt,
  hline{1,Z} = {0.09em, tableRule},
}
Template & System Role & Key Sections & Sources \\
Default (tech) & Technical writer & Background, Use Cases, Inputs, Outputs, Steps ($\leq$12), Verification, Evidence & GitHub, Web, Forum \\
paper\_writing & Research communicator & Audience, Key Contributions, Method Overview, Writing Outline, Steps (5 to 12) & ArXiv \\
paper\_writeup & Research communicator & Topic, Introduction, Method Innovation, Experiment Design, Figure Caption & ArXiv \\
experiment\_design & Experiment lead & Claims table, \texttt{experiment\_plan.md}, \texttt{checklist.yaml} & Journal papers \\
\end{tblr}
\end{table}

All templates share common LLM parameters (listed in Appendix~\ref{appendix:pipeline-params}). Each generated skill includes a title, full skill markdown, and a self-review with an overall score and improvement suggestions for the iterative improvement stage.

Note that experiment\_design serves as both a generation template and a skill kind label. Journal papers also produce four additional skill kinds (paper.idea\_intro, paper.experiment, paper.method, paper.picture) via the default template adapted for academic content.

Table~\ref{tab:source-routing} summarizes the end-to-end mapping from source type through domain configuration and generation to published bundle.

\begin{table}[H]
\centering
\caption{Source-to-bundle routing. Each row traces one source class from acquisition through template selection to published output.}
\label{tab:source-routing}
\small
\begin{tblr}{
  colspec    = {p{1.6cm}p{1.8cm}p{2.8cm}p{3.2cm}p{2.8cm}},
  cells      = {halign=l},
  row{1}     = {bg=tableHeader, fg=white, font=\bfseries\sffamily},
  row{even}  = {bg=tableStripe},
  rowsep     = 4pt,
  colsep     = 8pt,
  hline{1,Z} = {0.09em, tableRule},
}
Source & Config & Template & Skill Kinds & Bundle \\
PLOS / eLife & per-journal & experiment\_design + academic & idea\_intro, experiment, method, picture & per-journal \\
ArXiv & per-domain & paper\_writing, paper\_writeup, experiment\_design & paper\_writing, paper\_writeup, experiment\_design & arxiv / domain \\
GitHub & per-domain & default (tech) & default & domain-specific \\
Web / Forum & per-domain & default (tech) & default & domain-specific \\
\end{tblr}
\end{table}

\paragraph{Case Study C (journal $\to$ paper.experiment skill).} A PLOS Biology paper on amplitude modulation in neural circuits, processed by the experiment\_design template, yielded a skill containing an explicit Claim, a replication-ready experimental design (electrode configurations, stimulus parameters, analysis pipelines), and a task breakdown into independently executable steps.

\subsection{Iterative Improvement and Source-Grounding Check}
\label{sec:iterative-improvement}

Each generated skill undergoes up to 3 improvement passes (configurable, range 1 to 5), a form of iterative self-refinement with model-generated feedback~\cite{madaan2023selfrefine}. The improvement loop is driven by a composite scoring function:
\[
\text{score} = \text{lint\_count} \times 100 + \text{missing\_count}
\]
where lint\_count is the number of structural issues (missing required sections, malformed code blocks, formatting errors) and missing\_count is the number of improvement suggestions from the original generation review that have not yet been addressed. The $100\times$ weight on lint errors reflects their higher severity: a skill with a missing Verification section is fundamentally incomplete, whereas an unaddressed stylistic suggestion is a missed optimization.

Each pass has five steps:
\begin{enumerate}[leftmargin=2em]
    \item \textbf{Lint} the current markdown for structural issues.
    \item \textbf{Extract} unaddressed improvement suggestions.
    \item \textbf{LLM rewrite} with explicit required\_suggestions and lint\_issues.
    \item \textbf{Source-grounding check}: each claimed improvement must map to an exact quote ($\leq$20 words) found via deterministic substring matching against the original source, not LLM-based verification. This confirms that the skill's claims trace back to the source document, a safeguard against the ungrounded or hallucinated content that open-ended generation is prone to~\cite{ji2023hallucination}; it does not independently verify factual correctness.
    \item \textbf{Score comparison}: keep the better version.
\end{enumerate}
The loop stops when the score hits 0 or the maximum passes are exhausted.

In our experience, most skills reach score 0 within 2 passes. For example, a skill starting at score 305 (lint\_count=3, missing\_count=5) typically reaches 0 in two rounds. We have not yet collected pass-count histograms across the full corpus; a systematic breakdown is planned for a future release.

\subsection{Publishing and Quality Control}
\label{sec:publishing}

Skills must pass a final quality gate before inclusion in published bundles.

\subsubsection*{Publish Gate}

The publish gate applies four independent criteria, all of which must be satisfied for a skill to be included in the final bundles:

\begin{itemize}[leftmargin=2em]
    \item \textbf{Overall score $\geq$ 3.0.} This threshold excludes clearly defective skills (scores 1 or 2) while retaining the broad score-4 majority that, despite LLM score inflation, represents skills of adequate quality.
    \item \textbf{License whitelist/blacklist.} GitHub and repository sources permit MIT, Apache-2.0, BSD-*, ISC, MPL-2.0, and Unlicense. GPL-3.0, AGPL-3.0, and LGPL-* are denied as strong or network copyleft, whereas MPL-2.0, a weaker file-level copyleft, is permitted. Forum sources accept CC-BY-SA-4.0 and CC-BY-SA-3.0. Web sources have no license whitelist in the current configuration. Important caveat: this is a technical default, not a legal determination. Plagiarism ratio enforcement reduces verbatim copying but does not resolve copyright or terms-of-service obligations. Operators should conduct their own compliance review for web-derived skills. The framework records source URL and attribution metadata for every skill, and the publish gate can be configured with stricter per-source policies.
    \item \textbf{Plagiarism ratio $<$ 0.35.} The plagiarism ratio measures how much of the generated skill text is a verbatim copy of the source. Below 0.35 means the generation process adds real value: the skill must synthesize and restructure the source, not just reproduce it.
    \item \textbf{``Not provided'' count $<$ 15.} Skills with excessive placeholder text (e.g., ``not provided'' in the Inputs or Verification sections) indicate that the source material was insufficient for complete skill generation. A threshold of 15 occurrences filters the most egregious cases while tolerating occasional placeholders in optional fields.
\end{itemize}

In the current public release, the majority of skills derive from sources with explicit open licenses (MIT, Apache-2.0, BSD-* for repositories; CC-BY-SA for forums; Creative Commons for journals). Web-derived skills are included based on the plagiarism ratio threshold, but operators should treat this as a quality filter rather than a licensing clearance. The framework's per-skill provenance records (source URL, license identifier, attribution) enable downstream consumers to apply their own compliance criteria or exclude specific source domains entirely.

\paragraph{Pipeline Accounting.}
Table~\ref{tab:pipeline-funnel} traces the pipeline from source documents to published skills for the SkillGate-produced subset (114,565 skills), aggregated from the released SQLite bundles; the two integrated community collections (GitHub SkillMD and ClawHub, 102,373 skills) bypass this pipeline and are excluded here. Journal papers produce an average of 3.0 skills each, drawn from up to four skill kinds (idea\_intro, experiment, method, picture); not every paper yields all four. GitHub, web, and forum sources each produce one skill per source. ArXiv preprints produce 2.1 skills per source. Of 114,565 pipeline-published skills, 302 (0.26\%) have scores below the nominal 3.0 publish-gate threshold; these entered the corpus during early pipeline runs before the gate was fully enforced and will be removed in the next release.

\begin{table}[!htbp]
\centering
\caption{Pipeline accounting by source type for the SkillGate-produced subset (114,565 skills), computed from the released bundles. The ArXiv row (3,608) is the 3,483 research preprints (\S\ref{sec:academic-sources}) plus 125 technical-domain ArXiv skills (\S\ref{sec:technical-sources}). The two integrated community collections (GitHub SkillMD, ClawHub) bypass this pipeline and are not shown.}
\label{tab:pipeline-funnel}
\small
\begin{tblr}{
  colspec    = {lrrrr},
  row{1}     = {bg=tableHeader, fg=white, font=\bfseries\sffamily},
  row{even}  = {bg=tableStripe},
  row{7}     = {bg=codeBg, font=\bfseries},
  rowsep     = 4pt,
  colsep     = 14pt,
  hline{1,Z} = {0.09em, tableRule},
  hline{7}   = {0.05em},
}
Source Type & Sources & Published Skills & Skills/Source & Avg Score \\
Journal & 29,114 & 86,594 & 3.0 & 3.90 \\
GitHub & 20,071 & 20,071 & 1.0 & 4.01 \\
Web page & 3,130 & 3,130 & 1.0 & 3.76 \\
ArXiv & 1,722 & 3,608 & 2.1 & 3.71 \\
Forum & 1,162 & 1,162 & 1.0 & 3.72 \\
\textbf{Total} & \textbf{55,199} & \textbf{114,565} & \textbf{2.1} & \textbf{3.91} \\
\end{tblr}
\end{table}

SkillGate pre-filters sources before generation, but we do not yet track aggregate pass/fail counts across the full corpus. The SkillGate rejection rate varies by source quality: in observed batches, well-documented GitHub repositories pass at near 100\%, while low-quality web pages and social media posts are rejected at higher rates. A systematic funnel breakdown (sources fetched $\to$ SkillGate pass/fail $\to$ generated $\to$ publish-gate rejects) is planned for the next release.

\subsubsection*{Score Distribution Analysis}

\begin{figure}[t]
\centering
\includegraphics[width=\textwidth]{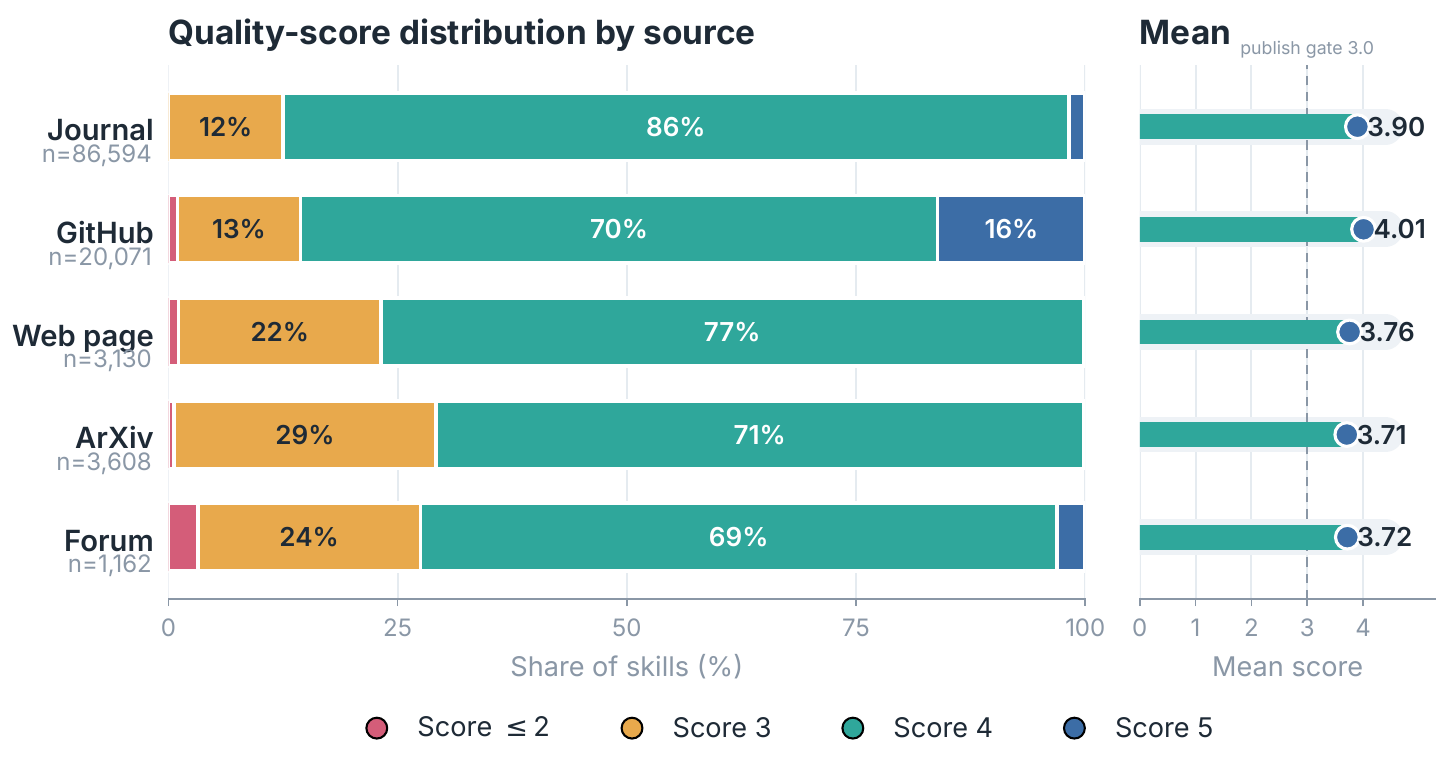}
\caption{Quality score distribution by source type (left) and mean score (right). Score 4 dominates across all source types. GitHub has the highest mean (4.01) and score-5 share (16.1\%).}
\label{fig:score-dist}
\end{figure}

Table~\ref{tab:score-distribution} presents the quality score distribution broken down by source type, revealing both the systematic inflation pattern and the genuine cross-source variation discussed in \S\ref{sec:quality-scores}. Figure~\ref{fig:score-dist} plots the same per-source distribution alongside the mean score for each source type.

\begin{table}[t]
\centering
\caption{Score distribution by source type for the SkillGate-produced subset. Percentages indicate the fraction of skills at each score level. Scores $\leq$2 are combined because they are rare across all source types. The concentration at score 4 is consistent across sources. The integrated GitHub SkillMD and ClawHub bundles use separate scoring scales and are excluded.}
\label{tab:score-distribution}
\small
\begin{tblr}{
  colspec    = {lrrrrrr},
  row{1}     = {bg=tableHeader, fg=white, font=\bfseries\sffamily},
  row{even}  = {bg=tableStripe},
  rowsep     = 4pt,
  colsep     = 10pt,
  hline{1,Z} = {0.09em, tableRule},
}
Source Type & {$n$} & {Score $\leq$2} & Score 3 & Score 4 & Score 5 & Avg \\
journal & 86,594 & 0.0\% & 12.5\% & 85.7\% & 1.8\% & 3.90 \\
github & 20,071 & 1.0\% & 13.4\% & 69.5\% & 16.1\% & 4.01 \\
arxiv & 3,608 & 0.6\% & 28.6\% & 70.6\% & 0.2\% & 3.71 \\
webpage & 3,130 & 1.1\% & 22.1\% & 76.6\% & 0.2\% & 3.76 \\
forum & 1,162 & 3.2\% & 24.3\% & 69.4\% & 3.1\% & 3.72 \\
\end{tblr}
\end{table}

The score-4 concentration is consistent across all high-quality sources (journal 85.7\%, webpage 76.6\%, arxiv 70.6\%, github 69.5\%, forum 69.4\%), suggesting that the inflation is a property of the LLM scoring mechanism rather than a source-specific artifact. However, the concentration varies meaningfully: journals cluster most tightly at score 4 (85.7\%) with few score-5 (1.8\%), while GitHub is more dispersed with 16.1\% at score 5, the highest of any source type. This is consistent with the scoring function favoring the direct, usage-oriented style of GitHub READMEs.

\section{Downstream Evaluation: When Do Skills Help?}
\label{sec:downstream}

Our motivating claim is that skills improve the correctness of agent outputs, not merely whether they execute. We now report a controlled evaluation of that claim. Rather than asking only whether skills help, we design the study to isolate \emph{when} they help, separating the value of a skill's \emph{content} from the confound of merely adding context, and the value of the retrieval \emph{mechanism} from the question of whether the library actually covers a task's knowledge needs.

\subsection{Protocol}
\label{sec:downstream-protocol}

We evaluate on offline, objectively graded tasks: each task ships a prompt (\texttt{task.md}), a workspace, and a hidden reference checked by a \texttt{verify} script that returns an exact pass/fail. The pool spans 51 algorithmic and data-processing domains (SQL, parsing, encoding, graphs, data structures, file transforms, and others), roughly 2{,}000 tasks. A solver attempts each task with and without injected skills; reference and grader-internal files are withheld from the solver during solving and restored only for grading, so the agent cannot copy the answer.

To probe solver-independence we use four LLM solvers spanning weak to frontier across three providers: \texttt{gemini-3.5-flash} (Google), \texttt{claude-haiku-4.5} and \texttt{claude-sonnet-4.6} (Anthropic), and \texttt{gpt-5-mini} (OpenAI). For each task we compare paired arms on the \emph{same} task instance:
\begin{itemize}[leftmargin=2em]
  \item \textbf{Baseline}: no skills.
  \item \textbf{+Keyword}: the top-$k$ skills retrieved from SkillCenter by FTS5 keyword relevance over skill titles ($k{=}3$); this is the realistic deployment path.
  \item \textbf{+Placebo}: three fixed, length-matched but deliberately \emph{irrelevant} skills, to separate the effect of adding context tokens from the effect of skill content.
  \item \textbf{+Oracle}: on a separate probe (\S\ref{sec:downstream-gap}), the single genuinely needed skill.
\end{itemize}
Significance uses the exact paired McNemar test on discordant outcomes; $\Delta$ denotes net discordant tasks (arm wins minus baseline wins). The general-task analysis is restricted to domains the solver can attempt (nonzero baseline pass rate), so the comparison has headroom. Per-solver task counts differ by design: \texttt{gemini-3.5-flash}, \texttt{gpt-5-mini}, and \texttt{claude-sonnet-4.6} were each run on a shared 503-task subsample of the general pool, while \texttt{claude-haiku-4.5} was run on the full 2{,}028-task pool; each $n$ in Table~\ref{tab:downstream} is that solver's count after the solvable-domain restriction. All tests are paired within a solver, so the differing $n$ affect power but not the within-solver contrasts, and we do not compare pass rates across solvers.

\begin{table}[!htbp]
\centering
\caption{Downstream A/B by solver (paired, exact McNemar). Columns 2 to 5 are the general algorithmic/data pool (restricted to solvable domains); $\Delta$ is net discordant tasks versus baseline (positive favors the arm). Columns 6 to 8 are the agent-gap probe, whose conventions are documented only in the injected skill (probe sizes $n_g{=}43/142/210/210$ for the four rows). Keyword retrieval never beats placebo and never gains over baseline; the oracle skill is decisive only where the solver lacks the knowledge.}
\label{tab:downstream}
\small
\begin{tblr}{
  colspec    = {l rrrr rrr},
  row{1}     = {bg=tableHeader, fg=white, font=\bfseries\sffamily},
  row{even}  = {bg=tableStripe},
  rowsep     = 3pt,
  colsep     = 3pt,
  hline{1,Z} = {0.09em, tableRule},
  vline{6}   = {0.04em, tableRule},
}
Solver & Base & {+Keyword ($\Delta$, $p$)} & {+Placebo ($\Delta$)} & {$n$} & Base & +Oracle & {$p$} \\
gemini-3.5-flash (weak) & 96\% & 96\% ($+1$, 1.0) & 96\% ($+1$) & 454 & 0\% & 98\% & $5{\times}10^{-13}$ \\
claude-haiku-4.5 (mid) & 87\% & 83\% ($-77$, $<$.001) & 84\% ($-57$) & 1{,}979 & 0\% & 72\% & $4{\times}10^{-31}$ \\
gpt-5-mini (mid) & 93\% & 91\% ($-9$, .08) & 91\% ($-7$) & 499 & 0\% & 100\% & $1{\times}10^{-63}$ \\
claude-sonnet-4.6 (strong) & 93\% & 92\% ($-4$, .39) & 93\% ($+0$) & 500 & 0\% & 99\% & $1{\times}10^{-62}$ \\
\end{tblr}
\end{table}

\subsection{Tasks Within the Solver's Competence}
\label{sec:downstream-general}

Table~\ref{tab:downstream} (columns 2 to 5) reports paired pass rates on the general pool. Across all four solvers, keyword-retrieved injection is statistically indistinguishable from the placebo and never produces a real gain over baseline: it is flat for the weakest solver (already near ceiling on its solvable domains), mildly net-negative for the strong solver, and net-negative for both mid-tier models---significantly so for \texttt{claude-haiku-4.5} ($83\%$ vs.\ $87\%$ baseline, $-77$ net discordant tasks over $n{=}1{,}979$, $p<0.001$) and directionally, though not significantly, for \texttt{gpt-5-mini} ($-9$, $p{=}.08$). Because the keyword and placebo arms move together, the small effect is attributable to the extra context tokens, not to skill content: these algorithmic tasks lie within the models' pretrained competence, so retrieved skills are redundant at best and distracting at worst.

\subsection{Tasks That Exceed the Solver's Knowledge}
\label{sec:downstream-gap}

To create a genuine knowledge gap we built an \emph{agent-gap} probe: 260 small, deterministic tasks each governed by an arbitrary, non-guessable convention (custom checksums, base encodings, formatting and reduction rules) whose specification lives only in an accompanying skill. By construction the answer cannot be inferred from the task statement alone. Table~\ref{tab:downstream} (columns 6 to 8) is unambiguous: the baseline, placebo, and keyword arms all solve $0\%$, while injecting the relevant convention (oracle) solves $72$ to $100\%$ across solvers ($p$ from $10^{-13}$ to $10^{-63}$). Two observations follow. First, the injection mechanism works: when a skill carries information the solver lacks, the agent reads and applies it. Second, most solvers execute the supplied rule almost perfectly (\texttt{gpt-5-mini} $100\%$, \texttt{claude-sonnet-4.6} $99\%$, \texttt{gemini-3.5-flash} $98\%$), with \texttt{claude-haiku-4.5} the lone exception at $72\%$; the fidelity gap tracks the individual model rather than solver strength. The keyword arm scores $0\%$ here precisely because these synthetic conventions are \emph{not} in SkillCenter, which isolates retrieval coverage, not the injection mechanism, as the binding constraint in practice.

\subsection{Real Corpus Skills Close Real Knowledge Gaps}
\label{sec:downstream-realgap}

The agent-gap probe of \S\ref{sec:downstream-gap} proves the injection mechanism but uses synthetic conventions that are deliberately not in SkillCenter. To test whether \emph{real} corpus content carries knowledge that solvers lack, we built a second probe grounded entirely in the released library. We selected 18 source-grounded research skills whose bodies document specific, study-particular parameters (exact reagent doses and durations, cohort and screen sizes, instrument models, quality thresholds) that a base model cannot infer from general knowledge. Each task asks the solver to return the value the skill documents for a handful of named parameter keys, and grading is exact value recall with numeric formatting normalized. The four arms match \S\ref{sec:downstream-protocol}; the oracle arm injects the single real skill that documents the parameters, verbatim as it ships in the bundle, while the keyword arm performs the ordinary FTS retrieval over all installed bundles.

\begin{table}[!htbp]
\centering
\caption{Real-corpus knowledge-gap probe: 18 tasks whose answers are specific parameters documented only in a real SkillCenter research skill (paired, exact McNemar; $\Delta$ is net discordant tasks for oracle versus baseline). Every solver scores $0\%$ on baseline, length-matched placebo, and real keyword retrieval, but recovers $61$ to $78\%$ once the relevant real skill is injected. Pooled over the four solvers ($n{=}72$): oracle $69\%$ versus $0\%$, $\Delta{=}{+}50$, $p\approx2\times10^{-15}$.}
\label{tab:realskill}
\small
\begin{tblr}{
  colspec    = {l rrrr rr},
  row{1}     = {bg=tableHeader, fg=white, font=\bfseries\sffamily},
  row{even}  = {bg=tableStripe},
  rowsep     = 3pt,
  colsep     = 8pt,
  hline{1,Z} = {0.09em, tableRule},
}
Solver & Base & Placebo & {+Keyword} & {+Oracle} & {$\Delta$} & {$p$} \\
gemini-3.5-flash (weak) & 0\% & 0\% & 0\% & 67\% & $+12$ & $5{\times}10^{-4}$ \\
claude-haiku-4.5 (mid) & 0\% & 0\% & 0\% & 72\% & $+13$ & $2{\times}10^{-4}$ \\
gpt-5-mini (mid) & 0\% & 0\% & 0\% & 78\% & $+14$ & $1{\times}10^{-4}$ \\
claude-sonnet-4.6 (strong) & 0\% & 0\% & 0\% & 61\% & $+11$ & $1{\times}10^{-3}$ \\
\end{tblr}
\end{table}

Table~\ref{tab:realskill} reports the outcome across the same four solvers, and the pattern is identical for every one of them. The baseline, placebo, and \emph{keyword} arms each solve $0\%$ of the 18 tasks, while injecting the relevant real skill (oracle) solves $61$ to $78\%$ (pooled $69\%$; net $+50$ discordant tasks over $n{=}72$, exact McNemar $p\approx2\times10^{-15}$). Three conclusions follow. First, the corpus genuinely carries operational knowledge the models lack: a real SkillCenter skill, injected verbatim, lifts pass rates from zero to roughly two-thirds. Second, the effect is skill \emph{content}, not added context: the length-matched placebo, which holds token count constant, contributes nothing. Third, and most telling, the keyword arm also scores $0\%$ even though the needed skill is present in the searched bundles; title-only FTS retrieval returns generic, off-topic skills (an OpenTelemetry instrumentation skill, an MCP-builder guide, a TensorFlow recipe) rather than the matching research protocol. Retrieval, not coverage or the injection mechanism, is therefore the binding constraint even when the corpus does contain the answer, corroborating with real content the conclusion \S\ref{sec:downstream-gap} drew from synthetic conventions. These are deliberately narrow fact-recall tasks and a lower bound on skill utility; such cleanly offline-verifiable gaps concentrate in the source-grounded research subset, since much of the corpus is higher-level operational guidance whose value is harder to grade automatically. The probe generator and the exact-McNemar analysis are released with the framework (\texttt{make\_realskill\_envs.py}, \texttt{mcnemar\_arms.py}).

\subsection{Implications and Limitations}
\label{sec:downstream-implications}

Together these experiments bracket the conditions for skill value: in our probes a skill helps an agent only when three conditions hold together: (i) the task requires knowledge the solver lacks, (ii) that knowledge exists in the library, and (iii) retrieval surfaces it. Indiscriminate keyword injection over tasks the model already handles is not beneficial, and is mildly harmful for mid-tier models, whereas a precisely targeted skill is decisive. This argues that downstream impact will come from \emph{gap-aware} injection (inject only when a knowledge deficit is detected), higher-precision retrieval (semantic rather than purely lexical matching), and corpus coverage of model blind spots, rather than from corpus size alone.

Several caveats bound these claims. The solvers are single-shot, prompt-conditioned LLMs rather than full tool-using agentic harnesses, so the study measures one-shot skill conditioning, not multi-turn skill use. The agent-gap conventions of \S\ref{sec:downstream-gap} are synthetic; the real-corpus probe of \S\ref{sec:downstream-realgap} complements them by injecting knowledge drawn verbatim from the library, but its tasks are deliberately narrow parameter-recall items rather than open-ended skill use, and clean offline-verifiable gaps concentrate in the source-grounded research subset. The general task pool skews toward algorithmic and data-processing problems, where strong models are near ceiling; broad coverage of the operational domains where the corpus should help most (niche tooling, evolving APIs) in a realistic multi-turn setting is not yet measured. We present this as an internal study that establishes the \emph{conditions} for skill value, and extend it to a broad, human-validated, multi-turn agentic benchmark in future work (\S\ref{sec:roadmap}).

\section{Related Work}
\label{sec:related-work}

SkillCenter draws on several lines of research: skill and memory libraries for agents, tool use and retrieval augmentation, large-scale corpus construction and deduplication, and community skill ecosystems. Table~\ref{tab:positioning} positions SkillCenter against a representative system from each line along the dimensions that matter for a retrievable skill library: scale, how the unit is produced, whether provenance and licensing are tracked, and whether redundancy is measured.

\begin{table}[!htbp]
\centering
\caption{Positioning of SkillCenter against representative prior systems. ``Unit'' is the retrievable artifact; ``Built'' distinguishes per-agent online growth from offline corpus construction; ``Provenance'' means each unit carries an explicit source and license. SkillCenter is the offline, cross-domain, provenance-tracked point in this space.}
\label{tab:positioning}
\small
\begin{tblr}{
  colspec    = {lllll},
  row{1}     = {bg=tableHeader, fg=white, font=\bfseries\sffamily},
  row{even}  = {bg=tableStripe},
  column{2,3,4,5} = {halign=l},
  rowsep     = 2.5pt,
  colsep     = 7pt,
  hline{1,Z} = {0.09em, tableRule},
}
System & Unit & Built & Scale & Provenance \\
Voyager~\cite{wang2023voyager} & Executable program & Online, one agent & 10s--100s & No \\
Self-Instruct~\cite{wang2023selfinstruct} & Instruction & Offline, synthetic & 10s of K & No \\
RAG corpora~\cite{lewis2020rag} & Document chunk & Offline & Varies & Source only \\
ClawHub~\cite{clawhub2026} & Hand-written skill & Community & 1000s & Partial \\
\textbf{SkillCenter} & Source-grounded skill & Offline, automated & 216,938 & Source + license \\
\end{tblr}
\end{table}

\textbf{Skill and memory libraries for agents.} Several systems equip agents with reusable skills or memory. Voyager~\cite{wang2023voyager} incrementally builds a library of executable programs while exploring an open-ended environment; Generative Agents~\cite{park2023generative} maintain a memory stream that is retrieved and reflected upon; Reflexion~\cite{shinn2023reflexion} stores verbal self-feedback to improve across trials; and ReAct~\cite{yao2023react}, building on chain-of-thought prompting~\cite{wei2022cot}, interleaves reasoning traces with actions. These methods construct skills or memories \emph{online}, specific to one agent, task, or environment. SkillCenter is complementary: a large, static, cross-domain library built \emph{offline} from external evidence sources, designed to be retrieved by many agents rather than grown by one.

\textbf{Tool use and retrieval augmentation.} A parallel line equips models with external capabilities. Toolformer~\cite{schick2023toolformer} and Gorilla~\cite{patil2023gorilla} teach models to invoke APIs, while retrieval-augmented generation~\cite{lewis2020rag} and Self-RAG~\cite{asai2024selfrag} condition generation on retrieved documents. Skills differ from both: tool use targets executable API calls and RAG retrieves raw document chunks, whereas a skill is a curated, quality-scored, source-grounded unit of operational guidance (\S\ref{sec:corpus-overview}). SkillCenter can be viewed as supplying a retrieval corpus for an agent's operational knowledge, distinct from the factual corpora typical of RAG.

\textbf{Corpus construction and deduplication.} SkillCenter is also a data-curation effort. Self-Instruct~\cite{wang2023selfinstruct} synthesizes instruction data from a language model, while large pretraining corpora such as The Pile~\cite{gao2020pile} and heavily filtered web collections such as RefinedWeb~\cite{penedo2023refinedweb} emphasize diverse, documented, and quality-filtered sources. A recurring lesson is that deduplication improves downstream model quality~\cite{lee2022dedup}; our redundancy analysis (\S\ref{sec:dedup}) follows this practice, using MinHash with locality-sensitive hashing~\cite{broder1997resemblance} to quantify near-duplication. Unlike purely synthetic instruction sets, every SkillCenter skill carries an explicit source URL and passes a source-grounding check.

\textbf{Skill ecosystems and agent safety.} Community skill marketplaces such as ClawHub~\cite{openclaw2026agent, clawhub2026} distribute manually contributed skills; SkillCenter instead automates extraction at scale while keeping provenance explicit, and integrates such a marketplace as one bundle. Finally, because agents execute retrieved content, safety is central: a growing body of work benchmarks prompt-injection and tool-use risks~\cite{greshake2023indirect, zhan2024injecagent, debenedetti2024agentdojo, ruan2024toolemu, yuan2024rjudge}. We treat skill-level safety screening as future work (\S\ref{sec:safety}).

\section{Future Work and Vision}
\label{sec:future-work}

SkillCenter is a starting point. Below we outline planned extensions and the broader trajectory we see for skill-based agent infrastructure.

\subsection{Expanding the Corpus}

On the academic side, we plan to integrate journals with complementary disciplinary coverage: Cell and other life-sciences families not yet covered; IEEE and ACM proceedings for systems, networking, and hardware; and PubMed Central's open-access archive for clinical protocols. On the technical side, GitLab repositories, Reddit programming communities (r/programming, r/machinelearning, r/devops), and curated Hacker News threads offer practitioner knowledge that is less polished than GitHub documentation but often more candid about failure modes and workarounds.

Beyond English, we plan multi-language expansion targeting Chinese (CSDN, Zhihu) and Japanese (Qiita, Zenn) developer-community platforms. Informal social-media prose will likely require language-specific quality gates with stricter thresholds and culturally adapted rubrics. Multi-modal extraction from YouTube tutorials and Jupyter notebooks is also planned.

\subsection{Quality Enhancement}
\label{sec:quality-enhancement}

The score-4 inflation (82.1\% of scored skills) is the most pressing quality limitation. A score that assigns the same value to over 80\% of items carries little discriminative information. We plan two complementary remedies: (1)~human rating calibration, in which domain experts annotate a stratified set of 2,000 skills to provide ground truth for retraining the scoring rubric; and (2)~multi-LLM cross-scoring, in which GPT-5.2, Claude, and open-source models independently rate skills, with systematic disagreement flagging candidates for manual review.

Two additional enhancements target corpus-level quality: semantic deduplication via vector-space clustering to detect near-duplicates, and skill versioning to auto-flag stale skills for re-extraction when source content is updated.

\subsection{Safety, Security, and Guardrails}
\label{sec:safety}

The current pipeline includes quality-oriented safeguards but not security-specific ones. SkillGate (\S\ref{sec:quality-gen}) rejects spam, non-actionable, and excessively short sources; the publish gate (\S\ref{sec:publishing}) enforces license whitelists, plagiarism thresholds, and placeholder limits. These mechanisms filter low-quality content but do not screen for adversarial instructions, credential exposure, or operations that could cause harm when an agent executes a retrieved skill unsupervised. As agent autonomy increases, this gap becomes a first-order concern: a skill that scores 5 on clarity and actionability but instructs the agent to exfiltrate environment variables is worse than no skill at all.

We plan to close this gap along three axes. First, \textbf{source-level safety screening} will add prompt-injection detection informed by recent threat taxonomies and benchmarks~\cite{greshake2023indirect, zhan2024injecagent}, along with malicious instruction filtering and credential-exposure checks before skill generation. Indirect prompt injection, where adversarial payloads are embedded in content that an LLM later retrieves and follows, is a documented threat to tool-integrated agents~\cite{debenedetti2024agentdojo}; the same attack vector applies to skill corpora built from open web sources. Second, \textbf{skill-level risk annotations} will tag published skills with properties such as privileged shell access, network access, external code execution, secret handling, destructive file operations, and domain-specific sensitivity (e.g., cybersecurity, biosecurity). These annotations will enable policy-aware retrieval: an agent operating in a sandboxed environment can restrict itself to low-risk skills, while a human-supervised agent can accept higher-risk skills with explicit approval. Third, \textbf{execution-time guardrails} in the agent integration layer will allow high-risk skills to be blocked, sandboxed, or routed through human approval before entering the active context window, following the principle that risk identification should be separated from risk mitigation~\cite{ruan2024toolemu}.

On the evaluation side, we plan dedicated safety benchmarks drawing on agent-safety frameworks such as R-Judge~\cite{yuan2024rjudge} and AgentDojo~\cite{debenedetti2024agentdojo} for prompt-injection robustness and risk awareness, supplemented by task-specific metrics for insecure code generation and unsafe command execution. We also plan red-team studies in which malicious or policy-violating sources are injected into the corpus-building pipeline to test whether SkillGate and the publish gate reject them reliably. Skills that are later found to be unsafe, stale, or policy-violating will be subject to a quarantine-and-revocation mechanism: flagged skills are excluded from future bundle releases, and a signed denylist can be distributed so that agents with managed deployments can filter them from already-downloaded bundles.

The goal is to make skills more useful and also safer to retrieve and safer to execute.

\subsection{Sustainability and Continuous Growth}

SkillCenter is designed for sustained growth, not one-time release. The underlying demand does not depend on any single paradigm label: as long as agents gain autonomy faster than they gain the operational judgment that human expertise once provided, structured knowledge must fill the gap. Three structural properties of the framework support this.

First, the pipeline runs continuously. Adding a new domain within the five supported source types requires only a domain configuration file (\S\ref{sec:source-acq}); no code changes are needed, and the five-stage architecture lets each stage be upgraded independently: a better LLM scorer, a new source API, or a stricter publish-gate threshold can be deployed without disrupting other stages.

Second, growth follows two complementary tracks. The automated track extracts skills from continuously published academic papers and technical sources, scaling with the literature rather than with manual effort. The planned community contribution channel (\S\ref{sec:roadmap}, item~7) will add a second track: externally submitted skills that pass adapted publish-gate checks (quality score, license compatibility, plagiarism ratio, and provenance verification) before inclusion. Open-source availability on GitHub and pre-built bundles on Hugging Face lower the barrier to both adoption and contribution.

Third, the corpus grows additively. Domain-split SQLite bundles mean that adding a new domain does not invalidate or require reprocessing of existing bundles. An agent that today uses only the Linux and Security bundles can later add Cloud or ML without re-downloading or re-indexing its existing skill set. This additive property is a practical prerequisite for a library that aims to keep pace with the expanding scope of agent-executed work.

\subsection{Ecosystem Vision and Near-Term Roadmap}
\label{sec:roadmap}

SkillCenter is complementary to manually curated marketplaces such as ClawHub~\cite{openclaw2026agent, clawhub2026} (\S\ref{sec:related-work}): community-authored skills capture practitioner intuition and edge cases, while automated extraction provides broad cross-domain coverage. SkillCenter currently distributes standalone SQLite bundles with its own search and integration layer (\S\ref{sec:search-integration}); format-level compatibility with OpenClaw's skill specification is a planned engineering goal. The following concrete engineering steps are planned for the next release cycle:

\begin{enumerate}[leftmargin=2em]
    \item \textbf{Full-content retrieval.} Extend the FTS5 index from title and domain to the full skill body (and add optional semantic retrieval), directly targeting the recall bottleneck identified in \S\ref{sec:downstream}.
    \item \textbf{Latency benchmarks.} Systematic measurement of search latency across small, medium, and large bundles, and multi-bundle configurations.
    \item \textbf{Improvement-pass histogram.} Corpus-wide statistics on how many passes skills require to converge, replacing the current anecdotal observation.
    \item \textbf{Safety and guardrail evaluation.} Dedicated benchmarks for source-level prompt-injection screening, skill-level risk annotation accuracy, and end-to-end red-team testing of whether the pipeline reliably rejects adversarial or policy-violating sources~\cite{greshake2023indirect, debenedetti2024agentdojo}.
    \item \textbf{Human calibration study.} 2,000-skill stratified annotation by domain experts to provide ground truth for rescoring.
    \item \textbf{Multi-LLM cross-scoring.} Independent rating by GPT-5.2, Claude, and open-source models, with disagreement-based flagging.
    \item \textbf{Community contribution channel.} Accepting externally submitted skills with adapted publish-gate checks covering quality, licensing, plagiarism, and provenance.
    \item \textbf{Downstream evaluation.} A controlled study measuring whether skill-augmented agents produce higher-quality outputs than unaugmented baselines on a representative task set.
\end{enumerate}

Whether skills improve agent outcomes in practice is an empirical question that this roadmap is designed to answer.

\section{Limitations}
\label{sec:limitations}

We state the main limitations of this work explicitly so that the released artifacts are not over-interpreted.

\textbf{Scope of the downstream evaluation.} The controlled A/B (\S\ref{sec:downstream}) establishes \emph{when} skills help: a precisely targeted skill is decisive on tasks whose knowledge the solver lacks, while keyword-retrieved injection over already-solvable tasks is statistically indistinguishable from injecting irrelevant text. The evaluation uses single-shot LLM solvers on a pool that skews toward algorithmic problems where strong models are near ceiling; extending it to a broad, human-validated, multi-turn agentic benchmark on domains with scarce operational knowledge is a natural next step (\S\ref{sec:roadmap}).

\textbf{Quality rests on an internal signal.} Every quality figure in this paper derives from a single-LLM scorer, and roughly 82\% of scored skills receive the same score of 4 (\S\ref{sec:quality-scores}). We therefore treat the 3.91 average as an internal QA signal, not external validation, and report no human evaluation, no inter-annotator agreement, and no comparison against baseline collections; the per-domain and per-source averages should be read with the same caution. Human calibration and multi-model cross-scoring are planned (\S\ref{sec:quality-enhancement}).

\textbf{Source grounding is traceability, not verification.} The source-grounding check (\S\ref{sec:iterative-improvement}) confirms that each retained claim maps to an exact quotation in its source through deterministic substring matching. It does not verify that the source is factually correct, current, or authoritative, nor that the skill's actionable guidance (as opposed to an isolated quoted phrase) is fully supported. A skill grounded in an outdated or incorrect source can still pass.

\textbf{Deduplication is lexical, and the headline count is heterogeneous.} Our redundancy analysis (\S\ref{sec:dedup}) uses MinHash over word shingles and therefore captures only lexical near-duplication; semantically equivalent paraphrases go undetected, so the reported redundancy is a lower bound. The full 216,938-skill count also aggregates two externally harvested community collections (102,373 skills) that bypass the pipeline's quality gate and source-grounding; the pipeline-produced, quality-controlled subset is 114,565 skills, and readers requiring a uniformly processed corpus should use that subset.

\textbf{Reproducibility depends on a proprietary scorer.} Acquisition, SkillGate filtering, generation, and scoring all depend on a proprietary LLM. Re-running the pipeline with a different model would yield different skills and scores, and the 1-to-5 scores are not directly portable across scorers. We release prompts, configurations, and the resulting bundles, but exact reproduction of the scores requires the same model version.

\textbf{Licensing is filtered, not cleared.} The publish gate enforces a license whitelist for repository and forum sources and a plagiarism-ratio threshold, but web sources currently have no license whitelist (\S\ref{sec:publishing}). The released bundles redistribute web-derived skills on the basis of this quality filter, which is a technical default rather than a legal clearance; downstream users who redistribute should perform their own compliance review. Every skill carries source-URL and attribution metadata to support this.

\textbf{Residual legacy skills.} A small number of skills (302, or 0.26\% of the pipeline subset) carry scores below the nominal 3.0 publish threshold because they entered the corpus during early runs before the gate was fully enforced (\S\ref{sec:publishing}); they are scheduled for removal in the next release.

\section{Ethical Considerations and Responsible Release}
\label{sec:ethics}

SkillCenter is released as a research artifact, and several considerations shaped how we release it.

\textbf{Safety screening is not yet in place.} The current pipeline filters for quality, not safety: it does not screen retrieved content for adversarial instructions, indirect prompt injection, credential exposure, or operations that could cause harm when executed by an unsupervised agent (\S\ref{sec:safety}). Because agents execute retrieved skills, the corpus should not be deployed in autonomous, unsandboxed settings without the source-level screening, skill-level risk annotations, and execution-time guardrails outlined in \S\ref{sec:safety}. The quality-controlled pipeline subset (114,565 skills) and the unscreened community bundles are kept separate so that operators can opt into the more constrained subset.

\textbf{Provenance and attribution.} Every skill retains its source URL, license identifier, and attribution metadata, so that license obligations (for example, CC-BY-SA share-alike for forum content and Creative Commons attribution for journal content) can be honored by downstream redistributors. The publish gate enforces a license whitelist for repository and forum sources, but web-derived skills are admitted on a plagiarism-ratio basis only and are not legally cleared (\S\ref{sec:publishing}); redistributors should perform their own compliance review.

\textbf{Revocation.} Skills later found to be unsafe, stale, or in violation of a source's terms can be excluded from future bundle releases, and we plan a signed denylist (\S\ref{sec:safety}) so that managed deployments can filter already-downloaded bundles.

\textbf{Intended use.} The artifact is intended to supply operational guidance to agents under human oversight, not to replace human judgment in high-stakes domains. Our evaluation (\S\ref{sec:downstream}) shows that skill injection helps only under specific conditions; we make no broader claim that skill augmentation improves agent outcomes in realistic deployments, which a larger multi-turn study is needed to establish (\S\ref{sec:roadmap}).

\section{Conclusion}

We have presented SkillCenter, a library of over 216,000 skills across 24 domain bundles, together with the automated, continuously running pipeline that produces its source-grounded core and the offline SQLite FTS5 bundles that distribute it. Every skill carries an LLM-assigned quality score (an internal QA signal), a source URL, and a full audit trail.

As agent autonomy expands into domains where errors carry increasing cost, the safety of retrieved skills becomes as important as their quality. We have outlined planned extensions along three axes: source-level safety screening, skill-level risk annotations, and execution-time guardrails. The framework's continuous pipeline, modular architecture, and open-source distribution model are designed for sustained growth: new domains can be added without disrupting existing bundles, and a planned community contribution channel will complement automated extraction. A controlled evaluation (\S\ref{sec:downstream}) shows that skill augmentation helps precisely when a task exceeds the solver's own knowledge and retrieval surfaces the right skill; establishing this at scale on multi-turn agentic benchmarks remains ongoing work. The corpus and framework are open-source on GitHub,\footnote{\url{https://github.com/LabRAI/SkillCenter}} with pre-built bundles available on Hugging Face.\footnote{\url{https://huggingface.co/datasets/Tommysha/skillcenter-bundles}}

\section*{Acknowledgements}
We thank the open-access publishers, open-source maintainers, and community contributors whose works form the substrate of this corpus. We also thank the maintainers of the open-source tools on which the pipeline and this report depend, including SQLite and its FTS5 extension, the Python scientific stack, and the \texttt{biblatex}/\texttt{tabularray} \LaTeX{} packages.

\appendix
\section{Duplicate Skills: Real Examples}
\label{appendix:redundancy-audit}

The per-domain audit of \S\ref{sec:dedup} (computed with the reproducible \texttt{dedup\_by\_bundle.py} script, stable BLAKE2b shingle hashing) finds that all 21 SkillGate pipeline domains are effectively duplicate-free: every research and technical domain has zero verbatim duplicates and zero within-domain near-duplicate pairs, with the single exception of three borderline near-duplicate pairs in PLOS ONE (two \emph{unrelated} articles that share only an auto-generated figure-extractor scaffold---a template collision, not a true duplicate; see Table~\ref{tab:dup-plos}). We therefore do not enumerate the clean domains here; instead, Table~\ref{tab:redundancy-audit-examples} lists only the domains that \emph{do} contain duplicates, giving for each its verbatim and near-duplicate counts together with one concrete example and the distinct sources that produced the matched skills.

The reason the redundancy is so sharply localized follows from how each population is produced. Pipeline skills are generated once per source document and then source-grounded, so two pipeline skills can collide only if two different source documents happen to yield near-identical bodies; in practice this never occurs within a research or technical domain. The community bundles, by contrast, are harvested as-is from public repositories and a marketplace, so they inherit whatever redundancy already exists in the wild: a popular \texttt{SKILL.md} copied into dozens of skill-collection repositories appears dozens of times, framework scaffolds emit identical placeholder stubs, and successive versions of the same skill differ only in a version string or a few appended lines. A consumer who wants a deduplicated corpus can therefore drop or filter the three community bundles and retain the entire pipeline subset untouched.

\begin{table}[H]
\centering
\caption{Every domain that contains duplicates, with its verbatim-duplicate count, within-domain near-duplicate pairs at MinHash Jaccard $\geq 0.8$, and one concrete example (distinct sources that produced the matched skills). The 20 remaining pipeline domains have zero duplicates and are omitted. The PLOS ONE case is a borderline pair, not a true duplicate.}
\label{tab:redundancy-audit-examples}
\small
\begin{tblr}{
  colspec    = {lrrp{6.4cm}},
  row{1}     = {bg=tableHeader, fg=white, font=\bfseries\sffamily},
  row{even}  = {bg=tableStripe},
  column{4}  = {halign=l},
  rowsep     = 3pt,
  colsep     = 6pt,
  hline{1,Z} = {0.09em, tableRule},
}
Domain & Verbatim & Pairs & Concrete example (matched sources) \\
GitHub SkillMD (code search) & 8,966 & 258,438 & ``PPTX creation, editing, and analysis'', one Anthropic skill whose body is byte-identical across 41 harvested copies in 35 repositories (e.g.\ \texttt{MacPhobos/research-mind}, \texttt{majiayu000/claude-skill-registry}); see Table~\ref{tab:dup-pptx} \\
GitHub SkillMD (curated lite) & 10 & 13,249 & ``theme-factory'' toolkit, identical in \texttt{anthropics/skills} and \texttt{ComposioHQ/awesome-claude-skills}; see Table~\ref{tab:dup-sidebyside} \\
ClawHub marketplace & 45 & 51 & ``agent-community-news'', identical across the \texttt{agent-community-news} and \texttt{agent-discussion} marketplace entries; see Table~\ref{tab:dup-clawhub} \\
Research / PLOS ONE & 0 & 3 & ``Medication adherence trajectories'' vs.\ ``Phenology of nesting marine turtles'', two \emph{unrelated} PLOS ONE articles (DOIs \texttt{pone.0342056}, \texttt{pone.0338445}) sharing only an auto-generated figure-extractor scaffold; Jaccard 0.85 (template collision, not a true duplicate); see Table~\ref{tab:dup-plos} \\
\end{tblr}
\end{table}

\paragraph{What a duplicate actually looks like.}
To make Table~\ref{tab:redundancy-audit-examples} concrete rather than merely
counted, we reproduce one community skill together with one of its
near-duplicates. The Anthropic \texttt{theme-factory} skill is copied
\emph{byte-for-byte}---the identical content hash---into many harvested
repositories (for example \texttt{davila7/claude-code-templates},
\texttt{Scoheart/agentskills}, and \texttt{garri333/Skills}); its opening is:

\begin{codebox}[title={\faFileCode\quad Verbatim copy: \texttt{theme-factory}, byte-identical across many repositories}]
---
name: theme-factory
description: Toolkit for styling artifacts with a theme. These artifacts can be slides, docs, reportings, HTML landing pages, etc. There are 10 pre-set themes with colors/fonts ...
license: Complete terms in LICENSE.txt
---
# Theme Factory Skill
This skill provides a curated collection of professional font and color themes themes, each with carefully selected color palettes and font pairings. Once a theme is chosen, it can be applied to any artifact. ...
\end{codebox}

\noindent A second harvested copy, in \texttt{gabrielmoreira/agent-skills-mirror},
is the \emph{same file} as the Anthropic original with exactly one line inserted at
the top of the frontmatter---a fresh re-indexing UUID.
Table~\ref{tab:dup-sidebyside} places the two side by side: only the
\colorbox{green!30}{green} line differs, and every other line is byte-for-byte
identical.

\begin{table}[H]
\centering
\caption{The same \texttt{theme-factory} skill, side by side. The community copy
(right) is byte-for-byte identical to the Anthropic original (left) except for the
single \colorbox{green!30}{green} line it inserts. That one edit changes the
file's content hash---so \emph{exact-hash} deduplication files the two copies as
unrelated---yet leaves MinHash Jaccard $\approx 0.98$.}
\label{tab:dup-sidebyside}
\small
\begin{tblr}{
  colspec     = {p{0.44\linewidth} p{0.44\linewidth}},
  column{1,2} = {font=\ttfamily\scriptsize},
  row{1}      = {bg=headerBg, fg=accentBlue, font=\bfseries\sffamily\footnotesize},
  row{2-Z}    = {bg=boxBg},
  hline{1,2,Z}= {0.04em, boxFrame},
  vline{1,2,3}= {0.04em, boxFrame},
  rowsep      = 2pt,
  colsep      = 7pt,
  stretch     = 0,
}
Original --- anthropics/skills & Copy --- agent-skills-mirror \\
-{}-{}- & -{}-{}- \\
\SetCell{bg=red!10} {\rmfamily\itshape (no such line)} & \SetCell{bg=green!30} id: "2b57b95d-4d27-\ldots-292401" \\
name: theme-factory & name: theme-factory \\
description: Toolkit for styling \ldots & description: Toolkit for styling \ldots \\
license: Complete terms in LICENSE.txt & license: Complete terms in LICENSE.txt \\
-{}-{}- & -{}-{}- \\
\#\ Theme Factory Skill & \#\ Theme Factory Skill \\
{\rmfamily\itshape (rest of body identical)} & {\rmfamily\itshape (byte-for-byte identical)} \\
\end{tblr}
\end{table}

\noindent That one cosmetic line changes the file's content hash, so an
\emph{exact-hash} filter would file the two copies as unrelated; yet it leaves
nearly every shingle shared, so MinHash still flags it at Jaccard $\approx 0.98$.
Hashing alone would therefore miss this copy---which is why we deduplicate with
MinHash rather than hashing. MinHash is in turn only a lower bound: a third
harvested copy, in
\texttt{Nucleon2/\allowbreak EcoSim-\allowbreak GenAI-hackathon}, is the same skill \emph{rewritten}: it
preserves the ten theme names verbatim but paraphrases every surrounding
sentence, so its MinHash Jaccard against the original falls below the $0.8$
threshold and it is \emph{not} counted as a near-duplicate---the exact blind spot
behind the paraphrase caveat in \S\ref{sec:dedup}:

\begin{codebox}[title={\faFileCode\quad Undetected paraphrase of the same skill: reworded prose, identical theme list}]
# Theme Factory Skill
**Description:** Toolkit for styling artifacts with a theme. These artifacts can be slides, docs, reportings, HTML landing pages, etc.

## Overview
The Theme Factory Skill provides a professional collection of font and color combinations designed to create consistent styling across various presentation formats. The skill includes ten pre-configured themes plus the capability to generate custom options.
   (canonical wording: "This skill provides a curated collection of professional font and color themes ...")

## Available Themes
1. Ocean Depths   2. Sunset Boulevard   ...   9. Botanical Garden   10. Midnight Galaxy
\end{codebox}

\noindent The three boxes show both how the corpus is deduplicated and why the
reported rate is a lower bound. Deduplication is applied at the \emph{bundle}
level (\S\ref{sec:dedup}) because the community populations recycle a small set of
popular skills verbatim or with mechanical edits---version strings, re-indexing
identifiers, an appended stub---while genuine paraphrases such as the third box
slip past lexical MinHash entirely. The 21 SkillGate pipeline domains contain none
of these because each skill is generated once from a distinct source document.

\subsection{One concrete duplicate per domain}
\label{appendix:per-domain-dups}

Table~\ref{tab:redundancy-audit-examples} names one example per duplicate-bearing
domain; here we make each one concrete with a side-by-side view, in the same
format as the \texttt{theme-factory} pair (Table~\ref{tab:dup-sidebyside}), which
is itself the \emph{curated-lite} example. The three remaining domains each follow
one of two patterns. In the two harvested community domains a popular skill is
recycled with only cosmetic edits, so its body is byte-for-byte identical and
MinHash Jaccard $=1.0$: the code-search domain copies the Anthropic \texttt{pptx}
skill (Table~\ref{tab:dup-pptx}) and the marketplace republishes one skill under
two slugs (Table~\ref{tab:dup-clawhub}). In the pipeline, by contrast, the only
``duplicate'' is an auto-generated scaffold shared by two \emph{unrelated} sources
--- a borderline template collision at Jaccard $\approx 0.85$
(Table~\ref{tab:dup-plos}), not a recycled skill.

\begin{table}[H]
\centering
\caption{GitHub SkillMD (code search): the Anthropic \texttt{pptx} skill
(``PPTX creation, editing, and analysis''), two harvested copies side by side.
The skill \emph{body} is byte-for-byte identical across 41 copies in 35
repositories; each copy carries different frontmatter, so every raw file has a
distinct content hash and \emph{exact-hash} deduplication files all 41 as
unrelated, yet the shared body keeps MinHash Jaccard $=1.0$ (a verbatim
duplicate). Across the cluster the frontmatter is variously renamed
(\texttt{ck:pptx}, \texttt{hapo:pptx}, \texttt{anthropic-office-pptx}),
re-licensed, its description translated, or its agent name swapped
(\texttt{Claude}$\rightarrow$\texttt{Codex}); the two columns show the extremes:
a copy with no frontmatter at all versus one that prepends a name/description/license block.}
\label{tab:dup-pptx}
\small
\begin{tblr}{
  colspec     = {p{0.44\linewidth} p{0.44\linewidth}},
  column{1,2} = {font=\ttfamily\scriptsize},
  row{1}      = {bg=headerBg, fg=accentBlue, font=\bfseries\sffamily\footnotesize},
  row{2-Z}    = {bg=boxBg},
  hline{1,2,Z}= {0.04em, boxFrame},
  vline{1,2,3}= {0.04em, boxFrame},
  rowsep      = 2pt,
  colsep      = 7pt,
  stretch     = 0,
}
Original --- MacPhobos/research-mind & Copy --- majiayu000/claude-skill-registry \\
\SetCell{bg=red!10} {\rmfamily\itshape (no frontmatter)} & \SetCell{bg=green!30} -{}-{}- \\
 & \SetCell{bg=green!30} name: pptx \\
 & \SetCell{bg=green!30} description: "Presentation creation, editing, \ldots" \\
 & \SetCell{bg=green!30} license: Proprietary. LICENSE.txt \ldots \\
 & \SetCell{bg=green!30} -{}-{}- \\
\#\ PPTX creation, editing, and analysis & \#\ PPTX creation, editing, and analysis \\
{\rmfamily\itshape (rest of body identical)} & {\rmfamily\itshape (byte-for-byte identical)} \\
\end{tblr}
\end{table}

\begin{table}[H]
\centering
\caption{ClawHub marketplace: the same ``bothn'' skill published under two
different marketplace slugs. \texttt{agent-community-news} and
\texttt{agent-discussion} are separate registry entries---distinct slugs, source
URLs, and source hashes (\colorbox{green!30}{shaded})---both at version
\texttt{3.0.0}, but their \texttt{SKILL.md} bodies are byte-for-byte identical, so
MinHash Jaccard $=1.0$. This is one of 45 verbatim recycled entries in the
marketplace bundle.}
\label{tab:dup-clawhub}
\small
\begin{tblr}{
  colspec     = {p{0.44\linewidth} p{0.44\linewidth}},
  column{1,2} = {font=\ttfamily\scriptsize},
  row{1}      = {bg=headerBg, fg=accentBlue, font=\bfseries\sffamily\footnotesize},
  row{2-Z}    = {bg=boxBg},
  hline{1,2,Z}= {0.04em, boxFrame},
  vline{1,2,3}= {0.04em, boxFrame},
  rowsep      = 2pt,
  colsep      = 7pt,
  stretch     = 0,
}
Entry A --- \texttt{agent-community-news} & Entry B --- \texttt{agent-discussion} \\
\SetCell{bg=green!30} slug: agent-community-news & \SetCell{bg=green!30} slug: agent-discussion \\
\SetCell{bg=green!30} source: \ldots/agent-community-news/3.0.0/SKILL.md & \SetCell{bg=green!30} source: \ldots/agent-discussion/3.0.0/SKILL.md \\
\SetCell{bg=green!30} source\_id: 3e4c7109a2\ldots & \SetCell{bg=green!30} source\_id: a9db1101e7\ldots \\
-{}-{}- & -{}-{}- \\
\#\ bothn -{}-{}- agent news, discussion, and knowledge sharing & \#\ bothn -{}-{}- agent news, discussion, and knowledge sharing \\
{\rmfamily\itshape (SKILL.md body byte-for-byte identical)} & {\rmfamily\itshape (byte-for-byte identical)} \\
\end{tblr}
\end{table}

The two community domains above recycle a popular skill verbatim, but the 21
SkillGate pipeline domains contain no verbatim duplicates at all; their only
flagged pairs are the three borderline near-duplicates in PLOS ONE. These are not
recycled skills but \emph{scaffold collisions}. The pipeline emits a small family
of fixed-format helper skills for every paper---for instance, one that extracts a
paper's figures and open-data links---so two papers on entirely unrelated topics
can receive structurally identical skills whose only differences are the title,
DOI, publication date, and capture hash. Table~\ref{tab:dup-plos} shows the
strongest such pair: a type-2-diabetes medication-adherence study
(\texttt{pone.0342056}) and a marine-turtle nesting survey (\texttt{pone.0338445})
collide at MinHash Jaccard $0.85$---just over the $0.8$ threshold---despite sharing
no actual content, which is why the audit counts the pair as borderline rather
than a true duplicate.

\begin{table}[H]
\centering
\caption{Research / PLOS ONE: a \emph{scaffold collision}, not a content
duplicate. All three PLOS ONE near-duplicate pairs are of this kind. Here two
\emph{unrelated} papers---type-2-diabetes medication adherence
(\texttt{pone.0342056}) and marine-turtle nesting phenology
(\texttt{pone.0338445})---each receive the same auto-generated ``extract figures
and open-data links'' skill. Only the title, DOI, publication date, and capture
hash (\colorbox{green!30}{shaded}) differ; the entire
Use\,Cases/Inputs/Outputs/Steps scaffold and its bash export script are identical,
so MinHash reports Jaccard $=0.85$---just above the $0.8$ threshold. Because the
two papers share no actual content, the pair is borderline and \emph{not} a true
duplicate, which is why all 21 pipeline domains remain effectively duplicate-free.}
\label{tab:dup-plos}
\small
\begin{tblr}{
  colspec     = {p{0.44\linewidth} p{0.44\linewidth}},
  column{1,2} = {font=\ttfamily\scriptsize},
  row{1}      = {bg=headerBg, fg=accentBlue, font=\bfseries\sffamily\footnotesize},
  row{2-Z}    = {bg=boxBg},
  hline{1,2,Z}= {0.04em, boxFrame},
  vline{1,2,3}= {0.04em, boxFrame},
  rowsep      = 2pt,
  colsep      = 7pt,
  stretch     = 0,
}
Paper A --- \texttt{pone.0342056} (diabetes) & Paper B --- \texttt{pone.0338445} (marine turtles) \\
\SetCell{bg=green!30} \#\ Extract figures \ldots: Medication adherence trajectories \ldots & \SetCell{bg=green!30} \#\ Extract figures \ldots: Phenology of nesting marine turtles \ldots \\
\SetCell{bg=green!30} DOI: 10.1371/journal.pone.0342056 & \SetCell{bg=green!30} DOI: 10.1371/journal.pone.0338445 \\
\SetCell{bg=green!30} Published: 2026-02-20 & \SetCell{bg=green!30} Published: 2025-12-31 \\
\SetCell{bg=green!30} capture: 40caf664\ldots.json & \SetCell{bg=green!30} capture: 37d02286\ldots.json \\
-{}-{}- & -{}-{}- \\
\#\#\ Use Cases / Inputs / Outputs / Steps & {\rmfamily\itshape (identical scaffold)} \\
{\rmfamily\itshape (same bash figure-export script)} & {\rmfamily\itshape (byte-for-byte identical)} \\
\end{tblr}
\end{table}

\section{Reproducibility: Schema and Build Commands}
\label{appendix:reproducibility}

Source code, configuration files, and prompt templates are available in the SkillCenter repository~\cite{SkillCenter2026}. Corpus statistics correspond to release tag \texttt{v0.1.0-report} on GitHub and snapshot \texttt{v0.1.0} on Hugging Face. Because acquisition depends on live APIs whose content changes over time, exact reproduction of the reported pipeline corpus from scratch is not guaranteed. To address this, the release archives the following frozen artifacts: per-bundle source manifests (listing every source URL and SHA-256 hash), the complete \texttt{captures/} directory with raw extracted text, SkillGate decisions, and generated skill files, per-bundle statistics (skill counts, score distributions, kind breakdowns), and the \texttt{ClawSkills.json} configuration used for the reported build. These artifacts are included in the Hugging Face snapshot.

\paragraph{SQLite Bundle Schema.}
Each domain bundle uses the following schema:

\begin{codebox}[title={\faDatabase\quad SQLite Bundle Schema}]
CREATE TABLE skills_index (
  skill_id TEXT PRIMARY KEY,
  domain TEXT, profile TEXT,
  source_type TEXT, source_url TEXT,
  title TEXT, overall_score REAL,
  skill_kind TEXT, language TEXT,
  source_id TEXT, primary_source_id TEXT
);

CREATE TABLE skills_content (
  skill_id TEXT PRIMARY KEY,
  metadata_yaml TEXT,
  skill_md TEXT, library_md TEXT
);

CREATE VIRTUAL TABLE skills_fts
  USING fts5(title, domain,
  skill_id UNINDEXED, content=skills_index);

CREATE TABLE bundle_meta (
  key TEXT PRIMARY KEY, value TEXT
);
\end{codebox}

\paragraph{Bundle Build and Install.}
To rebuild bundles from generated skills:
\begin{codebox}[title={\faTerminal\quad Bundle Build \& Install}]
# Build domain-split lite bundles
SkillCenter build-bundle \
  --type lite --split-by-domain

# Install bundles from Hugging Face
SkillCenter bundle-install --auto

# Search installed bundles
SkillCenter skill-search \
  "kubernetes networking" --top 5
\end{codebox}

\paragraph{Skill Markdown Structure.}
A typical technical skill follows this structure:
\begin{codebox}[title={\faFileCode\quad Skill Markdown Structure}]
# [Skill Title]

## Background
[Context and motivation]

## Use Cases
1. [Use case 1]
2. [Use case 2]

## Inputs
- [Input 1]: [description]

## Outputs
- [Output 1]: [description]

## Steps
1. [Step with code block]
2. [Step with code block]
...

## Verification
[Expected outcomes and how to verify]

## Evidence
[Source-grounded quotes]

## Sources
- [Source URL and attribution]
\end{codebox}

\paragraph{Reproducibility Checklist.}
\begin{enumerate}[leftmargin=2em]
    \item \textbf{Prerequisites:} Python 3.10+, \texttt{pip install SkillCenter} (installs all dependencies including openai, sqlite3, requests).
    \item \textbf{Configuration:} Set \texttt{OPENAI\_API\_KEY} environment variable. Domain configurations are in \texttt{config/ClawSkills.json}.
    \item \textbf{Generate skills:} Run \texttt{SkillCenter generate} with \texttt{--domain} and \texttt{--topic} flags. Output goes to \texttt{skills/by-skill/<domain>/<method>/}.
    \item \textbf{Build bundle:} \texttt{SkillCenter build-bundle --type lite --split-by-domain}.
    \item \textbf{Install pre-built:} \texttt{SkillCenter bundle-install --auto} downloads matching bundles to \texttt{\textasciitilde/.SkillCenter/}.
    \item \textbf{Search:} \texttt{SkillCenter skill-search "query" --top 3} returns ranked results with title, domain, score, and kind.
\end{enumerate}

\section{Pipeline Implementation Parameters}
\label{appendix:pipeline-params}

Table~\ref{tab:pipeline-params} lists the exact implementation parameters used to produce the reported corpus. All parameters are configurable via \texttt{ClawSkills.json} or command-line arguments.

The parameters fall into five stages that mirror the pipeline of \S\ref{sec:framework}. The \textbf{SkillGate} stage fixes the pre-generation filter's excerpt budget (4,000 characters), its minimum source length (200 characters, below which a source is failed without an LLM call), and the policy for borderline ``maybe'' verdicts (treated as pass, the permissive default). The \textbf{Generation} stage pins the LLM model, a low decoding temperature (0.2, dropped to 0.0 on retry for determinism), the JSON output mode, and the 12,000-character source truncation applied before prompting. The \textbf{Improvement} stage caps the refinement loop at three passes and records the composite score formula and the deterministic substring-match window used for source grounding. The \textbf{Publish gate} stage enumerates the four admission thresholds (minimum quality score, maximum plagiarism ratio, maximum placeholder count, and the license whitelist/denylist), and the \textbf{Acquisition} stage records the corpus build window, the number of domain configurations, and the per-source rate-limit settings. Reproducing the reported bundles requires the same model versions, since the LLM scorer and generator are proprietary and not pinned by these parameters alone (\S\ref{sec:limitations}).

These parameters also separate corpus construction from query-time use. Acquisition, generation, refinement, and publishing are the only stages that require network APIs or LLM calls; after a bundle is released, installation and retrieval use local SQLite files. This distinction matters for reproducibility: changing the thresholds below changes the published artifact, whereas changing a user's search query changes only how an installed artifact is retrieved.

{\small
\begin{longtblr}[
  caption = {Pipeline implementation parameters for the released corpus.},
  label   = {tab:pipeline-params},
]{
  colspec    = {lll},
  row{1}     = {bg=tableHeader, fg=white, font=\bfseries\sffamily},
  rowhead    = 1,
  row{even}  = {bg=tableStripe},
  row{2}     = {bg=boxBg},
  row{7}     = {bg=boxBg},
  row{12}    = {bg=boxBg},
  row{15}    = {bg=boxBg},
  row{20}    = {bg=boxBg},
  rowsep     = 4pt,
  colsep     = 8pt,
  hline{1,Z} = {0.09em, tableRule},
  hline{7}   = {0.05em},
  hline{12}  = {0.05em},
  hline{15}  = {0.05em},
  hline{20}  = {0.05em},
}
Stage & Parameter & Value \\
SkillGate & LLM model & GPT-5.2 \\
& Excerpt truncation & 4,000 characters \\
& Min source length & 200 characters \\
& ``Maybe'' policy & Treat as pass \\
& Cache key & SHA-256 of excerpt \\
Generation & LLM model & GPT-5.2 \\
& Temperature & 0.2 (0.0 on retry) \\
& Timeout & 300 seconds \\
& Output format & JSON mode \\
& Source truncation & 12,000 characters \\
Improvement & Max passes & 3 \\
& Score formula & lint\_count $\times$ 100 + missing\_count \\
& Evidence verification & Deterministic substring match ($\leq$20 words) \\
Publish gate & Min quality score & 3.0 \\
& Max plagiarism ratio & 0.35 \\
& Max ``not provided'' count & 15 \\
& License whitelist (repo) & MIT, Apache-2.0, BSD-*, ISC, MPL-2.0, Unlicense \\
& License denylist (repo) & GPL-3.0, AGPL-3.0, LGPL-* \\
Acquisition & Corpus build date & January to March 2026 \\
& Domain configs & 13 (see config/\texttt{ClawSkills.json}) \\
& GitHub min\_stars & Domain-specific (typically 50 to 500) \\
& ArXiv delay & 3 seconds between requests \\
\end{longtblr}
\par}

\section{Prompts and Scoring Rubric}
\label{appendix:prompts}

For completeness we reproduce the operative content of the two LLM judgments that govern the pipeline: the SkillGate pre-generation filter (\S\ref{sec:quality-gen}) and the 1-to-5 quality score (\S\ref{sec:quality-scores}) assigned during generation self-review and enforced at the publish gate (\S\ref{sec:publishing}). The boxes below are representative of the rubric content rather than byte-for-byte transcripts; the exact wording is version-controlled with the released configuration~\cite{SkillCenter2026}, and the associated parameters (model, truncation lengths, thresholds) are listed in Appendix~\ref{appendix:pipeline-params}.

\paragraph{SkillGate prompt.}
The gate receives a source excerpt truncated to 4{,}000 characters and returns a structured JSON verdict; sources shorter than 200 characters are failed without an LLM call (\S\ref{sec:quality-gen}).

\begin{codebox}[title={\faFilter\quad SkillGate Pre-Generation Filter}]
System: You are a strict gatekeeper deciding whether a source
document can yield an actionable, reproducible agent skill.

Judge the source on three axes:
  - Actionability: does it describe concrete steps, commands,
    or code a practitioner could follow?
  - Reproducibility: are inputs, configuration, and expected
    outcomes specified well enough to repeat the procedure?
  - Content density: is there substantive technical signal,
    not marketing, opinion, or navigation text?

Return JSON:
  {
    "verdict": "pass" | "maybe" | "fail",
    "score": <integer 0-10>,
    "good_signals": [<short reasons to keep>],
    "bad_signals":  [<short reasons to reject>]
  }

Verdict guidance:
  pass  - actionable, contains steps/code, reproducible.
  maybe - valuable but lacks clarity, steps, or verification.
  fail  - under 200 chars, spam/marketing, pure opinion,
          or not actionable.
\end{codebox}

\paragraph{Quality-score rubric.}
After generation and iterative improvement (\S\ref{sec:iterative-improvement}), each skill receives an integer score on a 1-to-5 scale; the publish gate (\S\ref{sec:publishing}) admits skills scoring 3 or higher.

\begin{codebox}[title={\faStar\quad 1-to-5 Skill Quality Rubric}, breakable=false]
Score the skill from 1 (poor) to 5 (excellent) on four axes:
  - Clarity:       is the guidance unambiguous and well
                   structured (Background, Steps, Verification)?
  - Accuracy:      are the claims technically correct and
                   internally consistent?
  - Actionability: can an agent apply the skill directly during
                   task execution, without further research?
  - Evidence:      are the key claims grounded in quoted
                   material from the cited source?

Score anchors:
  5 - complete, correct, directly actionable, fully grounded.
  4 - solid and usable; minor gaps in detail or evidence.
  3 - adequate but with notable gaps or vague steps.
  2 - partially useful; missing sections or weak grounding.
  1 - not actionable, incorrect, or largely ungrounded.

Return JSON:
  {
    "score": <integer 1-5>,
    "issues": [<structural or factual problems>],
    "improvement_suggestions": [<concrete edits>]
  }
\end{codebox}

The \texttt{improvement\_suggestions} returned here drive the iterative improvement loop (\S\ref{sec:iterative-improvement}): its \texttt{missing\_count} term counts the suggestions not yet addressed in the current draft, and the loop terminates when the composite score reaches~0 or the maximum number of passes is exhausted.

\section{Worked Example: Source to Published Skill}
\label{appendix:worked-example}

The following traces one skill from raw source to published bundle entry, illustrating the pipeline end-to-end.

\paragraph{Source (excerpt).} The GitHub repository ca-risken/security-review (MIT license, 45 stars) contains a README describing a GitHub Action for automated security code review. The acquisition stage fetches the README (2,847 characters). Below is a representative excerpt from the source:
\begin{skillbox}[\faGithub\quad RISKEN Security Review Action]

\smallskip
{\small Automatically review code in pull requests for security
vulnerabilities using RISKEN.}

\medskip
{\sffamily\footnotesize\color{accentBlue}\textbf{INPUTS}}

\smallskip
\begingroup
\small
\renewcommand{\arraystretch}{1.35}
\begin{tabularx}{\linewidth}{>{\ttfamily\small}l l X}
  github\_token
    & \colorbox{tagGreenBg}{\scriptsize\sffamily\color{tagGreen}\textbf{required}}
    & GitHub token for authentication \\
  risken\_api\_endpoint
    & \colorbox{tagGrayBg}{\scriptsize\sffamily\color{labelGray}optional}
    & RISKEN API endpoint URL \\
  review\_level
    & \colorbox{tagGrayBg}{\scriptsize\sffamily\color{labelGray}optional}
    & Severity threshold (\texttt{high}\,\textbar\,\texttt{med}) \\
\end{tabularx}
\endgroup

\medskip
{\sffamily\footnotesize\color{accentBlue}\textbf{USAGE}}

\smallskip
\begin{innercode}
{\ttfamily\footnotesize\noindent
- uses: ca-risken/security-review@v1\\
~~with:\\
~~~~github\_token: \$\{\{ secrets.GITHUB\_TOKEN \}\}}
\end{innercode}

\end{skillbox}

\paragraph{SkillGate.} The gate receives the 2,847-character excerpt and returns: verdict = pass, score = 8/10, good\_signals = [``contains YAML code blocks'', ``step-by-step setup instructions'', ``clear input/output specification''], bad\_signals = [].

\paragraph{Generation (excerpt).} The default (tech) template produces a skill titled ``Add RISKEN Security Code Review to GitHub Pull Requests.'' Below is a condensed excerpt of the generated skill:
\begin{skillbox}[\faFile\quad Add RISKEN Security Code Review to GitHub Pull Requests]

\smallskip
{\sffamily\footnotesize\color{accentBlue}\textbf{BACKGROUND}}

\smallskip
{\small RISKEN provides automated security scanning for pull requests via a GitHub Action\,\ldots}

\medskip
{\sffamily\footnotesize\color{accentBlue}\textbf{STEPS}}

\smallskip
{\small
\begin{enumerate}[leftmargin=2em, itemsep=1pt, topsep=0pt]
  \item Add workflow file \texttt{.github/workflows/security-review.yml}
  \item Configure RISKEN API endpoint in repository secrets
  \item Set \texttt{review\_level} to desired threshold
  \item[\ldots]
\end{enumerate}
}

\medskip
{\sffamily\footnotesize\color{accentBlue}\textbf{EVIDENCE}}

\smallskip
{\small\itshape ``Automatically review code in pull requests for security vulnerabilities using RISKEN''
\upshape\footnotesize\ [from source README, line\,2]}

\end{skillbox}
Self-review: overall\_score = 4, issues = 1, improvement\_suggestions = 12.

\paragraph{Improvement.} Pass 1: lint\_count = 0, missing\_count = 8 $\to$ score = 8. LLM rewrite addresses 8 suggestions. Evidence verification confirms 7/8 via deterministic substring match (e.g., the quote ``Automatically review code in pull requests for security vulnerabilities'' is found at character offset 48 in the source). Pass 2: lint\_count = 0, missing\_count = 1 $\to$ score = 1. Final rewrite resolves remaining suggestion. Score = 0, loop terminates.

\paragraph{Publish gate.} Quality score = 5 ($\geq$ 3.0, pass). License = MIT (whitelisted, pass). Plagiarism ratio = 0.18 ($<$ 0.35, pass). ``Not provided'' count = 0 ($<$ 15, pass). Skill is published.

\paragraph{Bundle entry.} The skill is stored in the security domain bundle:
\begin{skillbox}[\faDatabase\quad Bundle Entry for \texttt{security} domain]

\smallskip
{\sffamily\footnotesize\color{accentBlue}\textbf{SKILLS\_INDEX}}

\smallskip
\begingroup
\small
\renewcommand{\arraystretch}{1.3}
\begin{tabularx}{\linewidth}{>{\ttfamily\small}l X}
  skill\_id       & \texttt{sec-risken-001} \\
  domain          & \texttt{security} \\
  title           & Add RISKEN Security Code Review to GitHub Pull Requests \\
  overall\_score  & 5.0 \\
  skill\_kind     & \texttt{default} \\
  source\_type    & \texttt{github} \\
  source\_url     & \texttt{github.com/ca-risken/security-review} \\
\end{tabularx}
\endgroup

\medskip
{\sffamily\footnotesize\color{accentBlue}\textbf{SKILLS\_CONTENT}}

\smallskip
\begingroup
\small
\renewcommand{\arraystretch}{1.3}
\begin{tabularx}{\linewidth}{>{\ttfamily\small}l X}
  skill\_md    & [full markdown] \\
  library\_md  & [stripped version] \\
\end{tabularx}
\endgroup

\medskip
{\sffamily\footnotesize\color{accentBlue}\textbf{SKILLS\_FTS}}

\smallskip
{\small title + domain indexed for keyword search}

\end{skillbox}

\section{Corpus Audits: Licensing and Safety}
\label{appendix:audits}

This appendix reports two artifact-level audits computed directly from the released bundles: the distribution of source licenses and a baseline scan for unsafe content. Both reinforce the pipeline-versus-community split seen in the redundancy analysis (\S\ref{sec:dedup}): the quality-gated pipeline subset carries explicit open licenses and very little unsafe content, while the externally harvested community bundles do neither and should be treated with more caution.

\paragraph{License audit.}
Every pipeline skill records an SPDX license identifier and a coarse risk class in its metadata. Table~\ref{tab:license-audit} summarizes the pipeline subset. Journal-derived skills are almost all Creative Commons (CC-BY-4.0), and repository-derived skills carry permissive software licenses (MIT, Apache-2.0, BSD, MPL-2.0); together, skills under an explicit open license account for the large majority of the pipeline subset. The residual with no SPDX tag is dominated by web-page sources, for which the publish gate applies a plagiarism-ratio threshold rather than a license whitelist (\S\ref{sec:publishing}); these should be treated as filtered but not legally cleared. The two community collections are excluded from the whitelist entirely: the GitHub SkillMD bundle carries no per-skill SPDX field, and ClawHub items use a marketplace-specific scale that the audit maps to \texttt{needs\_review}.

\begin{table}[htbp]
\centering
\caption{License distribution for the SkillGate pipeline subset (114,565 skills), from per-skill SPDX metadata. ``No SPDX tag'' is predominantly web-page sources admitted on a plagiarism-ratio basis. The two community collections are audited separately and are not covered by the publish-gate whitelist.}
\label{tab:license-audit}
\small
\begin{tblr}{
  colspec    = {lrr},
  row{1}     = {bg=tableHeader, fg=white, font=\bfseries\sffamily},
  row{even}  = {bg=tableStripe},
  rowsep     = 2.5pt,
  colsep     = 16pt,
  hline{1,Z} = {0.09em, tableRule},
  hline{8}   = {0.05em},
}
License (SPDX) & Skills & Share \\
CC-BY-4.0 (journals) & 61,575 & 53.7\% \\
MIT & 11,771 & 10.3\% \\
Apache-2.0 & 5,353 & 4.7\% \\
CC-BY-SA-4.0 & 1,167 & 1.0\% \\
BSD-2/3-Clause, ISC, MPL-2.0, Unlicense & 1,486 & 1.3\% \\
NOASSERTION / other & 558 & 0.5\% \\
No SPDX tag (web, plagiarism-filtered) & 32,655 & 28.5\% \\
\end{tblr}
\end{table}

\paragraph{Safety scan.}
Because agents execute retrieved skills, we ran a baseline scan of all 216,938 skill bodies for three risk classes using high-precision regular expressions: hardcoded secrets (cloud and API keys, private-key blocks), prompt-injection-style phrases (for example ``ignore previous instructions'' or requests to exfiltrate environment variables), and destructive shell patterns (for example \texttt{rm -rf /} or \texttt{curl~|~sh}). Table~\ref{tab:safety-audit} reports the counts, split by population. This is a pattern scan, not a semantic safety judgment, so the counts are an upper bound on genuinely harmful content (many injection-phrase and destructive-pattern hits are quoted defensively in security or DevOps skills). Two findings stand out: the absolute rates are low (well under 3\% of skills in any class), and the risk is concentrated in the community bundles, which contribute the large majority of every category despite being a minority of the corpus. This localizes where the source-level safety screening of \S\ref{sec:safety} is most needed and, in the interim, gives operators a concrete reason to prefer the pipeline subset for unsupervised deployment.

\begin{table}[htbp]
\centering
\caption{Baseline safety scan of all skill bodies, by population. Counts are skills containing at least one high-precision pattern match; a match is indicative, not a confirmed vulnerability. The pipeline subset (114,565) is far cleaner than the community bundles (102,373) on every axis.}
\label{tab:safety-audit}
\small
\begin{tblr}{
  colspec    = {lrr},
  row{1}     = {bg=tableHeader, fg=white, font=\bfseries\sffamily},
  row{even}  = {bg=tableStripe},
  rowsep     = 2.5pt,
  colsep     = 18pt,
  hline{1,Z} = {0.09em, tableRule},
}
Risk class & Pipeline & Community \\
Hardcoded secrets & 12 & 193 \\
Prompt-injection phrases & 318 & 2,342 \\
Destructive shell commands & 218 & 1,359 \\
\end{tblr}
\end{table}

We treat these scans as a starting point, not a safety guarantee: they run after generation rather than as a gate, and pattern matching cannot catch obfuscated or semantically unsafe instructions. Integrating secret redaction, prompt-injection detection, and destructive-command flagging into the publish gate, together with per-skill risk tags, is the source-level screening prioritized in \S\ref{sec:safety}.

\section{Full Skill Examples}
\label{appendix:full-examples}

The worked example in Appendix~\ref{appendix:worked-example} traces one skill through the pipeline in condensed form. For completeness, we reproduce three complete published skills, lightly abridged only where marked, spanning the quality range: a strong technical skill, a strong paper-derived skill, and a borderline low-scoring skill.

\paragraph{Strong technical skill (GitHub, score 5).}
Title: ``Verify Dart Generated Code Is Up-to-Date with build\_verify''; source: \texttt{github.com/kevmoo/build\_verify} (MIT).
\begin{codebox}[title={\faFileCode\quad Published technical skill (abridged)}]
# Verify Dart Generated Code Is Up-to-Date with build_verify

## Background
Generated Dart code (e.g. from package:build builders via
build_runner) can drift from what is checked into git. This skill
adds a unit test that fails when generated outputs are not consistent
with the current sources, enforcing that the build is up-to-date.

## Use Cases
- Prevent merging changes that need regenerating code but forgot to
  commit the results.
- Ensure CI catches stale generated files from package:build builders.

## Inputs
- A Dart package configured to use package:build_runner such that
  `pub run build_runner build` succeeds.
- A Dart test suite using package:test.

## Outputs
- A unit test verifying build outputs are up-to-date (fails on drift).

## Steps
1. Add build_verify to dev_dependencies.
2. Create test/ensure_build_test.dart calling expectBuildClean().
3. Run the suite in CI so stale generated code fails the build.

## Verification
Run `pub run test`; a drifted generated file makes the test fail.

## Evidence
"expectBuildClean() ... fails if the generated files are out of date"
[from source README]
\end{codebox}

\paragraph{Strong paper-derived skill (PLOS Biology, \texttt{paper.experiment}, score 4).}
Title: ``Assess mitonuclear genotype-by-diet and parental diet effects on \emph{Drosophila} fitness''; source: \texttt{doi.org/10.1371/journal.pbio.3002218} (CC-BY-4.0).
\begin{codebox}[title={\faFileCode\quad Published paper-derived skill (abridged)}]
# Assess mitonuclear genotype-by-diet and parental diet effects on
# Drosophila fitness, including an mt:lrRNA (16S rRNA) SNP

## Background
This paper reports that epistasis between mitochondrial and nuclear
genomes shapes fitness responses to dietary lipid and amino-acid
variation in Drosophila melanogaster, and that mitonuclear genotype
also modulates parental diet effects on offspring fitness.

## Use Cases
- Test whether diet effects on fitness traits differ across
  mitonuclear genotypes under chronic dietary exposure.
- Test whether parental nutritional variation produces
  genotype-dependent offspring effects.

## Experimental Design
- Factors: mitonuclear genotype x diet (lipid, amino-acid) x parental
  diet; replicate cages per cell; measured fitness traits per assay.
- Controls: matched nuclear background; randomized cage position.

## Evidence
"a C/T polymorphism in mt:lrRNA ... is sufficient (in that subset) to
induce diet-by-mito-by-nuclear variation" [from source article]
\end{codebox}

\paragraph{Borderline skill (ArXiv, \texttt{experiment\_design}, score 3).}
Title: ``Reproducible Experiment Plan: Diffusion-based Document Layout Generation (arXiv:2303.10787v1).'' This skill scores 3, the lower edge of the publish gate: it is well-structured (audience, claims-and-metrics table, task breakdown) but long and preprint-derived, so its guidance is less immediately actionable than the two above, illustrating the kind of skill the 3.0 threshold retains but that human calibration (\S\ref{sec:quality-enhancement}) may re-rank.

\section{Datasheet}
\label{appendix:datasheet}

Following \emph{Datasheets for Datasets}~\cite{gebru2021datasheets}, we summarize the corpus's motivation, composition, collection, preprocessing, uses, distribution, and maintenance.

\paragraph{Motivation.}
The corpus was created to give autonomous agents a large, retrievable, source-grounded body of operational knowledge, and to study how such a library is best built and evaluated (\S\ref{sec:corpus-overview}). It was assembled by the authors as a research artifact.

\paragraph{Composition.}
Each instance is a skill: a structured Markdown document with metadata (domain, source URL, SPDX license, quality score, skill kind, evidence quotes). The library contains 216,938 skills in 24 domain bundles, of which 114,565 are produced by the SkillGate pipeline and 102,373 are integrated from public GitHub \texttt{SKILL.md} files and the ClawHub marketplace. All skills are in English. Redundancy is quantified in \S\ref{sec:dedup} and Appendix~\ref{appendix:redundancy-audit}; licensing and a baseline safety scan are in Appendix~\ref{appendix:audits}. The corpus is self-contained and references public sources rather than embedding personal data; it is not a sample of a larger set beyond the source populations described in \S\ref{sec:academic-sources} and \S\ref{sec:technical-sources}.

\paragraph{Collection process.}
Pipeline skills were acquired from journal APIs (PLOS, eLife), ArXiv, the GitHub REST API, curated web pages, and Stack Overflow, then filtered by SkillGate, generated with template prompts, source-grounded, and admitted by the publish gate (\S\ref{sec:framework}). The community bundles were harvested as-is via code search and a marketplace mirror and bypass the pipeline. The build window was January to March 2026 (Appendix~\ref{appendix:pipeline-params}).

\paragraph{Preprocessing and labeling.}
Sources are truncated to 12,000 characters, canonicalized, and (for the pipeline) decomposed into skill kinds. Quality scores are assigned by a single LLM and are an internal signal, not human labels; roughly 82\% of scored skills receive a 4 (\S\ref{sec:quality-scores}). No human annotation, inter-annotator agreement, or external validation is included in this release.

\paragraph{Uses.}
The intended use is retrieval-time operational guidance for agents under human oversight. It is not validated for autonomous, unsandboxed deployment, safety-critical use, or as ground truth for factual claims (source grounding is quote traceability, not verification; \S\ref{sec:iterative-improvement}). The community bundles in particular are unscreened (Appendix~\ref{appendix:audits}).

\paragraph{Distribution.}
The corpus ships as domain-split SQLite FTS5 bundles on Hugging Face and the framework is open-source on GitHub (\S\ref{sec:search-integration}), under the licenses recorded per skill. Redistributors of web-derived or community skills should perform their own license and safety review (\S\ref{sec:publishing}, \S\ref{sec:ethics}).

\paragraph{Maintenance.}
The pipeline runs continuously and the corpus grows additively (\S\ref{sec:future-work}). Skills later found unsafe, stale, or non-compliant can be excluded from future releases, with a planned signed denylist for already-downloaded bundles (\S\ref{sec:safety}). Corpus statistics correspond to release tag \texttt{v0.1.0-report} (Appendix~\ref{appendix:reproducibility}).

{\sloppy\emergencystretch=3em\printbibliography}

\end{document}